\documentclass[journal ]{new-aiaa}
\usepackage[utf8]{inputenc}
\usepackage{textcomp}
\usepackage[mathscr]{euscript}
\usepackage{graphicx}
\usepackage{amsmath}
\usepackage[version=4]{mhchem}
\usepackage{siunitx}
\usepackage{mathtools}
\usepackage{longtable,tabularx}
\usepackage{float}
\usepackage{bm}
\usepackage{vmargin}
\usepackage{booktabs, multirow} %
\usepackage{soul}
\usepackage{physics}
\usepackage{makecell}
\usepackage{subcaption} 
\usepackage{fancyhdr}
\usepackage{textcomp}
\usepackage{gensymb}
\usepackage{threeparttable}
\usepackage[table]{xcolor}
\usepackage{tablefootnote}
\usepackage{cleveref}
\usepackage{titlesec}
\usepackage{xspace}
\usepackage{textcomp}
\usepackage{etoolbox}
\usepackage{afterpage}
\usepackage{xurl}
\usepackage{layouts}
\usepackage{comment}
\crefname{equation}{Eq.}{Eqs.}
\crefname{table}{Table}{Tables}
\crefname{figure}{Fig.}{Figs.}
\crefname{section}{Sec.}{Secs.}
\crefname{appsec}{Appendix}{appendices}

\usepackage{textcase}
\DeclareMathAlphabet{\mymathbb}{U}{BOONDOX-ds}{m}{n}

\title{An Autonomous Vision-Based Algorithm for Interplanetary Navigation}

\author{Eleonora Andreis \footnote{Ph.D. Candidate, Department of Aerospace Science and Technology, Via La Masa, 34, eleonora.andreis@polimi.it} and Paolo Panicucci \footnote{Assistant Professor, Department of Aerospace Science and Technology, Via La Masa, 34, paolo.panicucci@polimi.it}} 
\affil{Politecnico di Milano, 20156, Milan, Italy}
\author{Francesco Topputo \footnote{Professor, Department of Aerospace Science and Technology,  Via La Masa, 34, francesco.topputo@polimi.it, AIAA Senior Member}}
\affil{Politecnico di Milano, 20156, Milan, Italy}

\begin{document}

\noindent{}This paper has been published in the Journal of Guidance Control and Dynamics.\newline
Please cite this paper as:\\[1cm]
{\ttfamily%
\renewcommand{\arraystretch}{0.8}
\begin{tabular}{p{1cm}p{15cm}}
@article\{andreis2024autonomous, & \\
&title = \{Autonomous Vision-Based Algorithm for Interplanetary Navigation\},\\
&author = \{Eleonora Andreis and Paolo Panicucci and Francesco Topputo\},\\
&journal = \{Journal of Guidance, Control, and Dynamics\},\\
&doi = \{10.2514/1.G007926\},\\
&year = \{2024\};\\
\} & \\
\end{tabular}}
\newpage
\maketitle

\begin{abstract}
The surge of deep-space probes makes it unsustainable to navigate them with standard radiometric tracking. Self-driving interplanetary satellites represent a solution to this problem. In this work, a full vision-based navigation algorithm is built by combining an orbit determination method with an image processing pipeline suitable for interplanetary transfers of autonomous platforms.
To increase the computational efficiency of the algorithm, a non-dimensional extended Kalman filter is selected as state estimator, fed by the positions of the planets extracted from deep-space images. An enhancement of the estimation accuracy is performed by applying an optimal strategy to select the best pair of planets to track. Moreover, a novel analytical measurement model for deep-space navigation is developed providing a first-order approximation of the light-aberration and light-time effects. Algorithm performance is tested on a high-fidelity, Earth--Mars interplanetary transfer, showing the algorithm applicability for deep-space navigation. %

\end{abstract}

\section{Introduction}
As a new era of deep-space exploration and exploitation is rapidly approaching, the adoption of efficient and sustainable navigation methods becomes increasingly crucial. Traditional ground-based radiometric tracking, while accurate and reliable, heavily depends on limited resources, such as ground stations and flight dynamics teams. This approach is unsustainable in the long term. There is then an urgent need to enhance the level of navigation autonomy for future interplanetary missions.

Different alternatives grant autonomous navigation capabilities: autonomous X-ray pulsar-based navigation \cite{WANG201427,malgarini2023}, semi-autonomous radio-based navigation \cite{oneway}, and autonomous vision-based navigation (VBN) \cite{henry2023absolute,maass2020crater}. Among these, X-ray navigation requires large detectors and long integration times \cite{turan2022autonomous}. One-way radiometric tracking still relies on Earth-based infrastructure. Whereas, VBN is an economical and fully ground-independent solution: it enables determining the probe position by observing the movement of celestial bodies on optical images \cite{turan2022autonomous}. In addition, VBN is an approach compatible with all mission phases toward celestial bodies: cruise \cite{bhaskaran2000deep,franzese2021deep,andreis2022onboard, henry2023absolute}, mid-range \cite{MERISIO2023115180,panicucci2023shadow, panicucci2023vision, pugliatti2022data}, and close proximity \cite{leilah2022}, including landing \cite{maass2020crater,LEROY2001787}. Several VBN solutions for approach and close proximity have been already adopted by different missions, e.g., NASA's OSIRIS-REx \cite{norman2022autonomous}. Instead, VBN algorithms for interplanetary navigation have only gone through onboard testing without being directly used for probe operations. An example is the validation performed within the Deep-Space 1 (DS1) mission in 1998 \cite{bhaskaran2012autonomous}. Nevertheless, in recent years, there has been a growing interest in VBN solutions for deep-space exploration, applied in particular to CubeSats missions \cite{turan2022autonomous}. %
Worth mentioning is the Miniaturised Asteroid Remote Geophysical Observer (M-ARGO), which aims to execute an onboard autonomous navigation test during the interplanetary transfer  \cite{topputo2021envelop}.

In interplanetary VBN, previous research primarily focuses on implementing orbit determination (OD) algorithms to determine the probe state \cite{raymond2015interplanetary,franzese2021deep,andreis2022onboard,casini2022line,stastny2008autonomous}. In \cite{raymond2015interplanetary}, innovative angles-only Initial Orbit Determination algorithms are developed, whose output is then used within an extended Kalman filter (EKF) embedding light-effects corrections on the planet position in the measurement model. In \cite{franzese2021deep}, the feasibility of the M-ARGO autonomous deep-space navigation experiment is presented. In \cite{andreis2022onboard}, an OD algorithm suited to be deployed on a miniaturized processor is developed by studying the most promising EKF implementations for onboard applications. Although the above works elaborate on autonomous OD, there is less literature focusing on developing a fully integrated pipeline embedding an image-processing (IP) procedure for extracting information from deep-space images. In \cite{bhaskaran2020autonomous}, an IP technique to retrieve beacon information is qualitatively mentioned yet not implemented in a fully integrated simulation, and the effect of the measurement errors on the state estimation is not quantified through simulations. Whereas, \cite{bhaskaran2000deep} details the procedure adopted to process the deep-space images of DS1. Due to the long exposure time and high-speed slew rate of the mission, complex image patterns were produced for the point sources. Thus, to retrieve accurately the centroids of the bright objects and the beacon position in the image, computationally heavy multiple cross-correlations were applied, following the approach used for the Galileo mission \cite{vaughan1992optical}. In this work, an alternative and computationally lighter approach has been preferred based only on geometrical evaluations following the assumptions of having slower slew rates. %

This work develops an autonomous VBN algorithm intended for use during a deep-space transfer, where the estimation accuracy is improved by applying light-effect corrections and an optimal strategy to select the best pair of beacons to track. The contribution to the state-of-the-art is threefold. First, the extended Kalman filter adopted as OD algorithm \cite{andreis2022onboard} is integrated with an IP pipeline suited to deep-space navigation \cite{andreis2022robust}. The literature in \cite{andreis2022onboard,casini2022line} is extended by considering deep-space images as input. In this way, the measurements are the outcome of an IP procedure rather than a mere behavioral model, which yields a more realistic representation of the application case and a faithful reproduction of the state estimation error. Second, the VBN filter is developed for CubeSat applications, thus, particular attention has been paid to the computation capabilities of the navigation algorithm. Third, a novel analytical measurement model for deep-space navigation providing a first-order approximation of light-effects correction on beacon position is presented. The proposed model avoids correcting the raw camera measurement, so decoupling the spacecraft prediction from the process noise and prevents onboard optimization as in \cite{andreis2022onboard}. Moreover, light-aberration correction is also applied to the 
stars position, being the attitude determined from deep-space images.

The paper is structured as follows. In Sec. \ref{sec: problem_statement} the interplanetary navigation problem is described by paying particular attention to the definition of the optimal beacon selection method and the light-effects perturbations relevant in the deep-space environment. Sec.\ \ref{sec:planetObs} details the IP procedure to extract observations from deep-space images. In Sec. \ref{sec: NAV filter}, the developed VBN filter to be used during an interplanetary transfer is presented. Here, the dynamics and measurement models are described together with the chosen filtering scheme.
Eventually, the performance of the IP pipeline and of the VBN filter tested on an interplanetary high-fidelity ballistic trajectory is reported in Sec. \ref{sec: results}.

\section{Interplanetary Vision-Based Navigation Problem}
\label{sec: problem_statement}
\subsection{Problem Geometry}
A probe can determine its location by acquiring information from the observation of celestial bodies through optical sensors. Since celestial objects are unresolved in deep space, i.e., they fall within a single pixel, their line-of-sight (LoS) direction or pixel position is the only available information that can be used to estimate the probe state. When two LoS directions associated with different beacons are obtained simultaneously, the kinematic celestial triangulation problem can be solved \cite{raymond2015interplanetary,kinematic,henry2023absolute}. %

In this work, CubeSats applications are investigated. This brings us to enforce some constraints, which make the navigation problem even more challenging than for standard probes:
\begin{enumerate}
\item only one miniaturized optical sensor, e.g., star tracker or camera, is adopted;
\item only planets are tracked because of the limited performance of the optical sensor \cite{franzese2023celestial};
\end{enumerate}
However, note that the algorithm can be used also for larger spacecraft, despite the application of the paper.

Since the kinematic celestial triangulation problem requires at least two different synchronous observations to be solved, but it is improbable to detect several planets with one single instrument due to their space sparsity, the static celestial triangulation cannot be exploited for the CubeSat operational scenario. Therefore, dynamic estimators, e.g., Kalman filtering, are adopted as they can process asynchronous observations. %

\subsection{Optimal Planets Selection}
To reach the highest accuracy possible in the state estimation, the approach described in \cite{franzese2020optimal} is adopted to optimally select the planets to observe during the interplanetary transfer. The optimal planets pair is chosen among the observable ones by minimizing the figure of merit $\mathcal{J}$, which is the trace of the position error covariance matrix when considering perturbed LoS directions. It is defined as follows:
\begin{equation}
    \mathcal{J} = \sigma_{\textrm{str}}^2 \  \dfrac{ 1 + \cos \gamma^2}{\sin \gamma ^4} \ \bm d^\top \ \Big( ( \bm I_{3x3} - \hat{ \bm \rho}_i\hat {\bm \rho}_i^\top) + (\bm I_{3x3} - \hat{\bm  \rho}_j \hat{\bm \rho}_j^\top) \Big) \ \bm d
\label{eq:J}
\end{equation}
where $\hat{ \bm \rho}_i$ and $\hat{ \bm \rho}_j$ are the unitary LoS vectors to the $i$-esimal and $j$-esimal planets, respectively, $\sigma_{\rm str}$ is the standard deviation of the LoS angular error, and $\bm I_{3 x 3}$ is the three-by-three identity matrix. Whereas, $\bm d$ and $\gamma$ are defined as %
\begin{align}
    \bm d= \bm r_{i} - \bm r_{j} && \gamma = \acos (  \hat{ \bm \rho}_i^\top \hat{ \bm \rho}_j )
\end{align}
where $\bm r_i$ and $\bm r_j$ are the positions of the two planets, respectively. It is convenient to divide $\bm d$ by 1 AU to keep $\mathcal{J}$ non-dimensional.

The optimal planet pair is selected taking into account the planets observability, which is preliminarily assessed by evaluating the planet apparent magnitude and Solar Exclusion Angle (SEA). For more information refer to \cite{andreis2022onboard}. %

\subsection{Light-Effects Perturbations}
Another important aspect to consider for deep-space navigation is the impact of light effects, i.e., light time and light aberration \cite{raymond2015interplanetary}, on the observations used to estimate the spacecraft state. On one hand, the light-time effect is caused by the fact that, given the large distance involved in deep space and the finite speed of the light, the light detected at the camera is emitted from the target in the past. This brings the celestial object to be observed shifted with respect to its position in the instant of detection. %
The further the planet is from the spacecraft, the more significant the light-time effect is. On the other hand, the light-aberration effect is caused by the relative motion between the observer and the light source, and it becomes important when the spacecraft velocity is not negligible. Like the light-time effect, this causes a change in the planet position projection, which depends on the velocity intensity and direction relative to the LoS of the observed planet. 

These two effects shall be corrected in the filter to avoid systematic errors in the estimation of the spacecraft state. Previous works consider these effects by applying corrections only to the planet LoS directions \cite{raymond2015interplanetary,andreis2022onboard}. Instead, in this work, since also the probe attitude is determined from deep-space images, the light-effect corrections need to be applied also to the computed stars LoS directions to avoid the evaluation of a biased attitude value. However, since stars are assumed to be fixed with respect to the Solar System, only light aberration needs to be corrected.

\section{Image Processing Pipeline for Deep-Space Vision-Based Navigation}
\label{sec:planetObs}
\begin{figure}[h!]
\centering
\includegraphics[width =\textwidth]{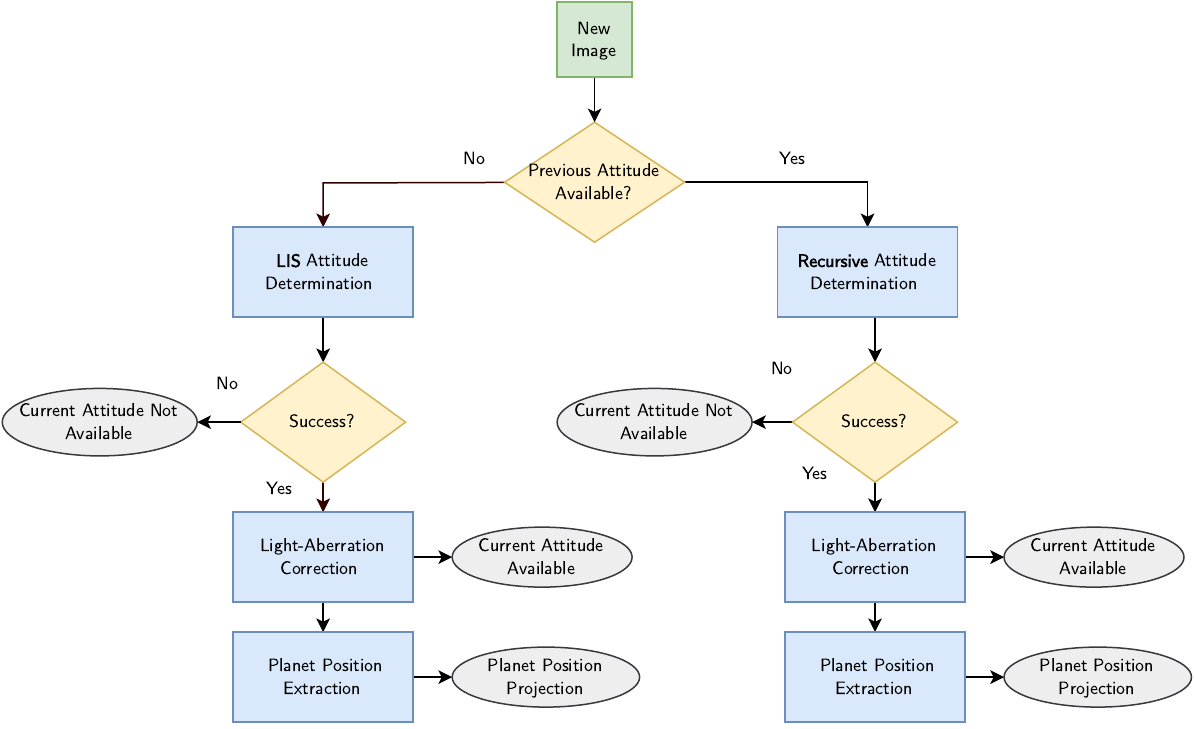}
\caption{Image Processing General Workflow}
\label{fig:acquisitionflow}
\end{figure}

In deep space, the projection of the planet position in the 2D camera reference frame $\mathbb{C}$, i.e., ${}_{}^{\mathbb{C}} \bm r_{\rm{pl}}$, or its associated LoS direction, is the only information available to support state estimation.
An IP algorithm suited for deep-space navigation is adopted to extract this information from the image. The goal of the IP procedure is to recognize the planet projections in the image among the centroids available.
The procedure goes through three steps: 1) The probe attitude is determined, 2) the light-aberration correction is applied to bright star centroids, and 3) the planets are identified. Note that the first step is needed to identify the portion of the sky the probe is observing and recognize those bright spots that correspond to not-stellar objects in the image. Although the current implementation foresees the attitude determination from the image, note that the Attitude Determination and Control System can also provide this solution in an operative scenario. A graphical representation of the IP procedure is shown in Fig.\ \ref{fig:acquisitionflow}. %
\subsection{Attitude Determination}
As the first step, the probe determines its attitude. At this aim, Niblack's thresholding method \cite{kazemi2015improving} is adopted to remove the background noise to portions of the image centered on bright pixels and delimited by squared windows with a margin of one pixel on each side. Hence, the centroid of the object is computed by applying an intensity-weighted center of gravity algorithm considering the pixels inside the associated squared window \cite{wan2018star}. At this point, the registration problem, whose goal is to find the correct matching between the observed star asterism and the cataloged stars in the inertial frame, is solved. This last step is performed differently according to whether the planet is acquired for the first time or not.

In the former case, the selected lost-in-space (LIS) strategy is the search-less algorithm (SLA) introduced in \cite{search-less}. In this work, the SLA has been preferred over the binary search technique \cite{bentley1997fast} for its higher speed gain rate (from 10 to more than 50 times \cite{mortari2014k}) and for its robustness to spikes. %
To be adopted, the SLA requires computing on the ground a vector of integers, the k-vector, which contains information for the stars matching starting from the chosen stars invariant. In this work, the interstellar angle is the invariant chosen to build the star catalog. To reduce the size of the catalog, only stars whose apparent magnitude is lower than 5.5 are considered for the generation of the invariant. Moreover, interstar angles greater than 35 deg are not taken into account.
The objects identified by the SLA as spikes may be non-stellar objects (such as planets, asteroids, and cosmic rays) or stars not recognized due to errors in the centroid extraction.
Yet, when a great number of spikes is present in the image, the star asterisms may not be recognized by the algorithm. In this work, to reduce the number of scenarios in which this failure occurs, a heuristic approach is considered. As faint stars are generally not stored in the onboard catalog and as the centroids extraction depends on the thresholding procedure,  when the attitude determination fails, the attitude determination procedure is iterated again by increasing the intensity threshold. One of the results of this approach consists of diminishing the number of bright objects in the image, which can ultimately lead to the removal of some spikes. The procedure is repeated until observed star asterisms are recognized or less than three stars are detected. 

When the spacecraft is not in LIS mode, it has a rough estimate of its orientation. Therefore, a recursive registration method can be applied. Indeed, by knowing the previous attitude estimation, the LoS directions in the inertial reference frame $\mathcal{N}$ of the four corners of the image are determined. At this point, a check is performed to identify which stars of the onboard catalog are contained inside the image Field of View (FoV). Thus, their position projections in the 2D camera reference frame are evaluated, and they are associated with the closest centroids of the bright objects extracted from the image.

When stars are identified, the probe attitude is determined by solving Wahba's problem \cite{markley2014fundamentals} between the stars' LoS directions in the camera and inertial reference frame exploiting the Singular Value Decomposition (SVD) method \cite{markley2014fundamentals}.
Moreover, the robustness of the solution to Wahba's problem is increased thanks to the adoption of a RANdom-SAmple Consensus (RANSAC) procedure \cite{fischler1981random,ransac}. The RANSAC algorithm aims to detect the bright objects that have been misidentified by the star identification, which can thus lead to a wrong attitude determination. %
To detect these outliers, the attitude of the spacecraft is adopted as the mathematical model for the data fitting. The attitude is estimated $n_R$-times by selecting randomly every time a group of 3 identified stars. The minimum set of stars needed for attitude determination is chosen to increase the probability of having a group made of different stars at each time. Thus, the estimated $n_R$ spacecraft orientations are compared to identify the best model, which is then adopted for the data fitting. The stars not respecting the best model are considered outliers and are labeled as spikes. %

When the recursive attitude determination fails, the spacecraft orientation at the following image acquisition is determined again with the LIS method. Vice versa, when the LIS algorithm succeeds in the determination of the probe orientation, in the following image acquisition the recursive attitude determination algorithm will be adopted. 
\subsection{Light-Aberration Correction}
\label{sec:light-aberration correction}
After the first attitude determination, the centroids of the stars are corrected for the light-aberration effect, and the probe attitude is recomputed by taking into account the corrected stars LoS directions. The procedure adopted is described in \cite{raymond2015interplanetary}. At first, the observed stars LoS directions as seen by the spacecraft in the inertial reference frame $\mathcal{N}$ are found:
\begin{equation}
    {}^{\mathcal{N}} \bm \rho_{s_{\rm{obs}}}  = (\bm K_{\rm{cam}} \bm A)^{-1}{}_{h}^{\mathbb{C}}\bm{r}_{s_{\rm{obs}}}
    \label{eq:2d23d}
\end{equation}
where ${}_{h}^{\mathbb{C}}\bm{r}_{s_{\rm{obs}}}$ are the observed stars position projections in $\mathbb{C}$ in homogeneous coordinates (see \cite{ransac} for homogeneous coordinates), and $\bm K_{\rm{cam}}$ is the camera calibration matrix. Then, the angle $\theta_{\textrm{obs}}$ between the observed unitary stars LoS directions ${}^{\mathcal{N}} \hat{\bm \rho}_{s_{\rm{obs}}}$ and the estimated unitary velocity vector of the probe $\hat{\bm v}_{\rm{p}}$ is defined as
\begin{equation}
    \tan{\theta_{\textrm{obs}}} = \frac{|| {}^{\mathcal{N}} \hat{\bm \rho}_{s_{\rm{obs}}} \times \hat{\bm v}_{\rm{p}}|| }{{}^{\mathcal{N}} \hat{\bm \rho}_{s_{\rm{obs}}}^\top \hat{\bm v}_{\rm{p}}}
\end{equation}
And the aberration angle $\varepsilon$ is evaluated:
\begin{equation}
    \tan{\varepsilon} = \dfrac{(v_{\rm{p}}/c) \sin\theta_{\textrm{obs}}}{1 - (v_{\rm{p}}/c) \cos \theta_{\textrm{obs}}}
\end{equation}
where $c$ is the speed of the light.
Thus, the corrected unitary stars LoS directions $ {}^{\mathcal{N}} \hat{\bm \rho}_{s_{\rm{corr}}}$ can be  retrieved as

\begin{equation} \label{eq: pl_ab_correction}
   {}^{\mathcal{N}}  \hat{\bm \rho}_{s_{\rm{corr}}} = \dfrac{{}^{\mathcal{N}}  \hat{\bm \rho}_{s_{\textrm{obs}}} \sin \theta_{\textrm{corr}} - \hat{\bm v}_{\rm{p}}\sin \varepsilon}{\sin{\theta_{\textrm{obs}}}}
\end{equation}
with $\theta_{\textrm{corr}} = \theta_{\textrm{obs}} + \varepsilon$.
At this point, the attitude matrix of the probe is redetermined by solving the Wahba problem \cite{markley2014fundamentals} in which $ {}^{\mathcal{N}}  \hat{\bm \rho}_{s_{\rm{corr}}}$ are considered. This corrected attitude matrix is labeled $\bm A_{\rm{corr}}$.
\subsection{Beacon Identification}
At this step, the planet must be identified in the image and its projection ${}_{}^{\mathbb{C}} \bm r_{\rm{pl}}$ must be extracted. In this step, the identification is performed through the evaluation of the statistical moments associated with the planet position projection, which defines the Gaussian probability of finding the planet in that portion of the image. At first, the expected position projection of the observed planet is evaluated as:
\begin{equation}
{}_{h}^{\mathbb{C}} \bm r_{\rm{pl}_0} = \bm K_{\rm{cam}} \bm A_{\rm{corr}} ({}_{}^{\mathcal{N}}\bm r_{\rm{pl}} - {}_{}^{\mathcal{N}}\bm r_{\rm{p}})
\end{equation}
where ${}_{}^{\mathcal{N}}\bm r_{\rm{p}}$ is the predicted probe position. 
If ${}_{}^{\mathbb{C}} \bm r_{\rm{pl}_0}$ falls within the boundaries of the image, its associated uncertainty ellipse is computed. The latter depends on the uncertainties of the spacecraft pose and planet position and is centered in ${}_{}^{\mathbb{C}} \bm r_{\rm{pl}_0}$. The ellipse of ${}_{}^{\mathbb{C}} \bm r_{\rm{pl}_0}$ represents the area of the image where the planet is most likely to be found within a 3$\sigma$ probability. The spike contained in the 3$\sigma$ ellipse is identified as the planet position projection ${}_{}^{\mathbb{C}} \bm r_{\rm{pl}}$. If multiple spikes are located within this ellipse, the closest one to the expected planet position is identified as the planet, as it is most likely to be the true planet projection. %

The covariance matrix of the beacon position projection $\bm P$ due to the spacecraft pose and beacon position uncertainty is computed as
\begin{equation}
     \bm P =\bm G \bm S \bm  G^\top
\end{equation}
where $\bm G$ is the Jacobian matrix of the mapping between the beacon position projection ${}^{\mathbb{C}}\bm r_{\rm{pl}}$ and the spacecraft pose and the beacon position, and $\bm S$ is the uncertainty covariance matrix of the probe pose and beacon position. 
To evaluate $\bm G$, the variation of $\ {}^{\mathbb{C}}\bm r_{\rm{pl}}$ with respect to the variation of the spacecraft pose and the beacon position is computed. In particular, the quaternions $\bm q= (q_0,\bm q_v)^\top$ are chosen to represent the probe attitude matrix. Eq. \eqref{eq:quaternions2dcm} gives the quaternion representation of the attitude matrix $\bm A_{\textrm{corr}}$ \cite{markley2014fundamentals}
\begin{equation}
    \bm A_{\textrm{corr}} = (q_0^2 - \bm q_v^\top\bm q_v) \bm I_{3x3} + 2\bm q_v\bm q_v^\top - 2 q_0 [\bm q_v]^\wedge
    \label{eq:quaternions2dcm}
\end{equation}
where $\left[{(\cdot)}\right]^{\wedge}$ is the skew-symmetric matrix associated with the cross-product operation.
Thus, the variation of $\ {}^{\mathbb{C}}\bm r_{\rm{pl}}$ with respect to the variation of the spacecraft pose, i.e., $\bm A_{\textrm{corr}}(\bm q_{\mathcal{C}/\mathcal{N}})$ and ${}_{}^{\mathcal{N}}\bm r$, and the beacon position ${}_{}^{\mathcal{N}}\bm r_{\rm{pl}}$ can be defined as
\begin{equation} 
 {\delta} \  {}^{\mathbb{C}}\bm r_{\rm{pl}} = \underbrace{\begin{bmatrix}
   \dfrac{\partial \ {}^{\mathbb{C}}\bm r_{\rm{pl}}}{\partial q_0} & \Big [\dfrac{\partial \ {}^{\mathbb{C}}\bm r_{\rm{pl}}}{\partial \bm q_v} \Big ]& \Big [  \dfrac{\partial \ {}^{\mathbb{C}}\bm r_{\rm{pl}}}{\partial {}_{}^{\mathcal{N}}\bm r} \Big ]  & \Big [\dfrac{\partial \ {}^{\mathbb{C}}\bm r_{\rm{pl}}}{\partial {}_{}^{\mathcal{N}}\bm r_{\rm{pl}}} \Big ] \\
    \end{bmatrix}}_{\bm G}
    \begin{pmatrix}
    \delta q_0 \\
    \delta \bm q_v \\
    \delta{}_{}^{\mathcal{N}} \bm r \\
    \delta {}_{}^{\mathcal{N}}\bm r_{\rm{pl}} \\
    \end{pmatrix}
\end{equation}
The matrix $\bm G$ has dimension $2\times10$ and it is defined as
\begin{equation}
  \bm G = {\begin{bmatrix}
     \dfrac{1}{{}_{h}^{\mathbb{C}}{ r}_{\rm{pl}_3}} & 0 & -\dfrac{{}_{h}^{\mathbb{C}}{r}_{\rm{pl}_1}}{{}_{h}^{\mathbb{C}}{ r}_{\rm{pl}_3}^2}\\
     0 & \dfrac{1}{{}_{h}^{\mathbb{C}}{ r}_{\rm{pl}_3}} & -\dfrac{{}_{h}^{\mathbb{C}}{r}_{\rm{pl}_2}}{{}_{h}^{\mathbb{C}}{ r}_{\rm{pl}_3}^2}
    \end{bmatrix}}\bm K_{\rm{cam}}
    \begin{bmatrix}
     \dfrac{\partial (\bm A_{\textrm{corr}} \ {}^{\mathcal{N}}{\bm \rho})}{\partial q_0} &  \dfrac{\partial (\bm A_{\textrm{corr}} \ {}^{\mathcal{N}}{\bm \rho})}{\partial \bm q_v}& \dfrac{\partial (\bm A_{\textrm{corr}} \ {}^{\mathcal{N}}{\bm \rho})}{\partial {}_{}^{\mathcal{N}}\bm r} & \dfrac{\partial (\bm A_{\textrm{corr}} \ {}^{\mathcal{N}}{\bm \rho})}{\partial {}_{}^{\mathcal{N}}\bm r_{\rm{pl}}}
    \end{bmatrix}
\end{equation}
where the partial derivatives of $(\bm A_{\textrm{corr}}\ {}^{\mathcal{N}}{\bm \rho})$ with respect to the spacecraft pose and beacon position are
\begin{equation}
     \frac{\partial (\bm A_{\textrm{corr}}\ {}^{\mathcal{N}}{\bm \rho})}{\partial q_0} =  2q_0{}^{\mathcal{N}}{\bm \rho} - 2[\bm q_v]^\wedge \ ^{\mathcal{N}}{\bm \rho}
\end{equation}
\begin{equation}
    \frac{\partial (\bm A_{\textrm{corr}} \ {}^{\mathcal{N}}{\bm \rho})}{\partial \bm q_v} = -2 \ {}^{\mathcal{N}}{\bm \rho}\bm q_v^\top + 2\bm q_v^\top \ {}^{\mathcal{N}}{\bm \rho}\bm I_{3x3} + 2\bm q_v \  {}^{\mathcal{N}}{\bm \rho^\top} + 2 q_0[{}^{\mathcal{N}}{\bm \rho}]^\wedge
\end{equation}
\begin{equation}
    \frac{\partial (\bm A_{\textrm{corr}} \ {}^{\mathcal{N}}{\bm \rho})}{\partial {}_{}^{\mathcal{N}}\bm r} = -\bm A_{\textrm{corr}}
\end{equation}
\begin{equation}
    \frac{\partial (\bm A_{\textrm{corr}} \ {}^{\mathcal{N}}{\bm \rho})}{\partial {}_{}^{\mathcal{N}}\bm r_{{\rm{pl}}}} = \bm A_{\textrm{corr}}
\end{equation}
A change of attitude representation is performed to define $\bm S$. Since the uncertainty of the probe orientation is more clearly identified through Euler's principal rotation theorem, the quaternion variation is linked to the one relative to the principal angle $\theta$, also known as pointing error, and principal axis $\bm e$. Moreover, a reference attitude value is considered to be always present onboard. Therefore, the variation with respect to the nominal value is limited. Thus, the small-error-angles formulation can be adopted \cite{markley2014fundamentals}:
\begin{align}
\delta q_0 = 0 && \delta \bm q_v = \frac{1}{2}\delta (\theta \bm e)
\end{align}
Since $\sigma_{\bm{q}_0}^2 = 0$ for the small-angles assumption, $\bm S$ can be described:
\begin{equation}
    \bm S = \rm{diag}(\sigma_{\bm{q}_v}^2{\bm I_{3x3}},\sigma_{\bm r}^2{\bm I_{3x3}},\sigma_{\bm r_{\rm{pl}}}^2{\bm I_{3x3}})
    \label{eq:varquat}
\end{equation}
where $\sigma_{\bm r}$ and $\sigma_{\bm r_{\rm{pl}}}$ represent the standard deviation of the probe position and beacon position, respectively. Note that the cross-correlations are ignored for simplicity, yet in a more integrated solution, the pose could be coupled.
Once the covariance matrix of the beacon position projection is assessed, the associated 3$\sigma$ uncertainty ellipse is computed. Let $\lambda_{\text{max}}$ and $\lambda_{\text{min}}$ be the largest and smallest eigenvalues of $\bm P$, respectively, and $\bm v_{\text{max}}$, $\bm v_{\text{min}}$ their related eigenvectors. Note that $\bm P$ has only two eigenvalues. %
The characteristics of the {3$\sigma$} covariance ellipse can be computed as:
\begin{align}
    a = \sqrt{11.8292 \ \lambda_{\text{max}}} && b = \sqrt{11.8292 \ \lambda_{\text{min}}} && \psi = \arctan{\Big (\frac{\bm v_{\rm{max}_2}}{\bm v_{\rm{max}_1}}\Big )}
\end{align}
where $a$ is the  {3$\sigma$ covariance ellipse} semimajor axis, $b$ the {3$\sigma$ covariance ellipse} semiminor axis, $\psi$ the {3$\sigma$ covariance ellipse} orientation (i.e., the angle of the largest eigenvector towards the image axis $\bm C_1$), and $\bm v_{\rm{max}_2}$, $\bm v_{\rm{max}_1}$ the eigenvector related to the maximum eigenvalue along $\bm C_2$ and $\bm C_1$ directions, respectively. Note that the value 11.8292 represents the inverse cumulative distribution function of the chi-square distribution with 2 degrees of freedom at the values in 0.9973 ($3\sigma$). %
Eventually, the beacon is identified with the closest spike to the expected beacon position projection contained in the 3$\sigma$ ellipse.  %
\section{Non-dimensional Extended Kalman Filter Based on Planets Observations }
\label{sec: NAV filter}
In this section, the VBN filter is described. Firstly, the dynamic and measurement models adopted in the VBN filter are detailed. Successively, the chosen filtering scheme is shown. Note that the vectors specified in this section are always defined in the inertial reference frame $\mathcal{N}$. Thus, the superscript is indicated only for exceptions.

\subsection{Dynamics Model}
The process state $\bm x$ is defined as
\begin{equation}
   \color{black} {\bm x(t) = [ {}^{\mathcal{}}\bm r(t), {}^{\mathcal{}}\bm v(t), \bm \eta(t)]^\top}
    \label{eq:state_repr}
\end{equation} \color{black}
where $ {}^{\mathcal{}}\bm r$ and $ {}^{\mathcal{}}\bm v$ are the inertial probe position and velocity, respectively, and $\bm \eta$ is a vector of Gauss--Markow (GM) processes accounting for unmodeled terms: a 3-dimensional residual accelerations $\bm \eta_{\textrm{R}}$ and the stochastic component of the Solar Radiation Pressure (SRP) $\bm \eta_{\textrm{SRP}}$; that is, $\bm \eta=[\bm \eta_{\textrm{R}}, \bm \eta_{\textrm{SRP}}]^\top$ \cite{bestpractise}. %
The process is modeled using the following equation of motion
\begin{equation}
    \dot{ \bm x}(t) = \bm f(\bm x(t),t) + \bm w
    \label{eq:system dynamics eq}
\end{equation}
where $\bm f$ is the vector field embedding the deterministic part, while $\bm w$ is the process white noise:
\begin{equation}
\dot{ \bm x}(t) =
     \underbrace{\begin{bmatrix}
    &&  {}^{\mathcal{}}\bm v&& \\
    &&  {}^{\mathcal{}}\bm a_{\rm{Sun}} +   {}^{\mathcal{}}\bm a_{\rm{SRP}} + \sum_{i }  {}^{\mathcal{}}\bm a_{\rm{pl}_i} &&\\
    && - \xi \bm \eta_{\textrm{R}}  && \\
    && - \xi \bm \eta_{\textrm{SRP}}  && \\
    \end{bmatrix}}_{\bm f}
    + 
    \underbrace{\begin{bmatrix}
    \label{eq:f}
    && \bm 0_{3x1} && \\
    &&  \bm \eta_{\textrm{R}} + \bm \eta_{\textrm{SRP}} && \\
    &&  \bm w_{\textrm{R}} && \\
    &&  \bm w_{\textrm{SRP}} && \\
    \end{bmatrix}}_{\bm w}
\end{equation}
and 
\begin{align}
 {}^{\mathcal{}}\bm a_{\rm{Sun}} = -\mu_{\textrm{Sun}}\dfrac{ {}^{\mathcal{}}\bm r}{|| {}^{\mathcal{}}\bm r||^3} \\
 {}^{\mathcal{}}\bm  a_{\rm{SRP}} = C_\textrm{R}\dfrac{P_\textrm{0}R_\textrm{0}^2}{c}\dfrac{A_\textrm{s}}{m_\textrm{s}}\dfrac{ {}^{\mathcal{}}\bm r}{|| {}^{\mathcal{}}\bm r||^3}\\
 {}^{\mathcal{}}\bm a_{\rm{pl}_i} = \mu_{\rm{pl}_i} \left( \frac{\bm  {}^{\mathcal{}} \bm 
r_{\rm{pl_i}} -  {}^{\mathcal{}}\bm r}{|| {}^{\mathcal{}} \bm r_{\rm{pl_i}} - {}^{\mathcal{}} \bm r||^3} - \frac{ {}^{\mathcal{}}\bm r_{\rm{pl_i}}}{|| {}^{\mathcal{}} \bm r_{\rm{pl_i}}||^3} \right)
\end{align}
The terms that describe the SRP are \cite{cannonball_model}: $C_\textrm{R}$ the coefficient of reflection, $P_\textrm{0}$ the solar power, $R_\textrm{0}$ the Sun radius, $A_\textrm{s}$ the cross-section area of the probe, and $m_\textrm{s}$ its mass. The third-body perturbations of the Earth-Moon barycenter, Mars, and Jupiter are included; $\mu_{\rm{pl}_i}$ is the gravitational parameter of the $i$-esimal planet considered.  In the Langevin equations that govern the GM processes the coefficient $\xi$ defines the reciprocal of the correlation time, while $\bm w_{\textrm{R}}$ and $\bm w_{\textrm{SRP}}$ are the process noises of the GM parameters with $\sigma_\textrm{R}$ and $\sigma_{\textrm{SRP}}$ standard deviations, respectively \cite{bestpractise}. The process noise covariance matrix is $\bm Q$:
\begin{equation}
\bm Q = \textrm{diag}(\bm 0_{3x3}, \bm Q_a, \bm Q_{\rm R}, \bm Q_{\rm SRP})
\end{equation}
with $\bm Q_\textrm{R} = \sigma_\textrm{R}^2 \bm I_{3x3}$, $\bm Q_{\textrm{SRP}} = \sigma_{\textrm{SRP}}^2 \bm I_{3x3}$, and $\bm Q_{\rm a} = (\bm Q_{\rm R} + \bm Q_{\rm SRP})/(2\xi)$.
\subsection{Measurement Model}
\label{sec:measurement model}
One of the contributions of the work is to present a novel measurement model for deep-space triangulation. In the state of the art the LoS measurements have been modeled as azimuth and elevation observations \cite{andreis2022onboard,raymond2015interplanetary,mortari2017single}. Thus, the raw camera measurement is manipulated to obtain a derived quantity to be included in the filter. This implies that the error distribution is less Gaussian with respect to the raw measurement error distribution. %
Moreover, the correction of the light-aberration effect is performed on the observation through the exploitation of the estimated velocity information, coupling the measurement and the process noise. 
On the contrary, the measurement model proposed hereafter expresses the observations in pixel coordinates in the camera plane. %
In addition, it embeds the light effects and their dependencies with respect to the planet and spacecraft state. Therefore, the navigation filter takes into account these effects during the mean and covariance update, and the coupling between the measurement and the process noise is avoided. 

\subsubsection{Evaluation of the time delay}
To proceed with the implementation of the light-time correction, it is first necessary to evaluate the time delay $\Delta t$. As the initial step, the equation representing the light-time effect can be written as follows \cite{andreis2022onboard}:
\begin{equation}\label{eq:constraintTau}
    \mathfrak{L} := c^2\left(t -\tau \right)^2 -  (\bm r_{\rm{pl}}(\tau) - \bm{r}(t))^{\top}(\bm r_{\rm{pl}}(\tau) - \bm{r}(t)) = 0
\end{equation}
where $\tau$ is the time at which the light is emitted by the planet and $t$ is the time of measurement. %

Eq.\ \eqref{eq:constraintTau} is a constraint that links the spacecraft state with the planet position. Moreover, as the planet motion does not have an analytical solution, the value of $\tau$ cannot be solved analytically to implicitly include this effect in the measurement. Yet, it is possible to linearize the constraint in Eq.\ \eqref{eq:constraintTau} with respect to $\Delta t$ under the assumption that the time delay $\Delta t = t - \tau$ is small.  The planet motion can be approximated linearly as:
\begin{equation}
    \bm{r}_{\rm{pl}}(\tau) \simeq \bm{r}_{\rm{pl}}(t) + \left. \frac{\text{d} \bm{r}_{\rm{pl}}}{\text{d} \tau} \right|_{\tau = t} \left( \tau - t  \right) = \bm{r}_{\rm{pl}}(t) - \left. \frac{\text{d} \bm{r}_{\rm{pl}}}{\text{d} \tau} \right|_{\tau = t} \Delta t = \bm{r}_{\rm{pl}}(t) - \bm{v}_{\rm{pl}}(t) \Delta t 
\end{equation}
Therefore, Eq.\ \eqref{eq:constraintTau} can be linearized as well as follows:
\begin{equation}
\begin{aligned}
    \mathfrak{L} &\simeq c^2\Delta t^2 + \left( \bm{r}_{\rm{pl}}(t) - \bm{v}_{\rm{pl}}(t)\Delta t -\bm{r}(t) \right)^\top\left( \bm{r}_{\rm{pl}}(t) - \bm{v}_{\rm{pl}}(t)\Delta t -\bm{r}(t) \right) =\\
    &= (c^2 - \bm{v}_{\rm{pl}}(t)^\top \bm{v}_{\rm{pl}}(t))\Delta t^2 +2\bm{v}_{\rm{pl}}(t)^\top \bm{r}_{\rm{pl/sc}}(t)\Delta t - \bm{r}_{\rm{pl/sc}}(t)^\top \bm{r}_{\rm{pl/sc}}(t)
\end{aligned}
\end{equation}
where $\bm{r}_{\rm{pl/sc}}(t) = \bm{r}_{\rm{pl}}(t) -\bm{r}(t)$. Thus, the solution $\Delta t$ is obtained by solving a second-order polynomial equation:
\begin{equation}\label{eq:twosolutions}
\begin{aligned}
\Delta t =&\frac{1}{c^2 - \bm{v}_{\rm{pl}}(t)^\top\bm{v}_{\rm{pl}}(t)}\left( -\bm{r}_{\rm{pl/sc}}(t)^\top \bm{v}_{\rm{pl}}^{}(t) \right.\\ &\left.\pm \sqrt{\bm{r}_{\rm{pl/sc}}(t)^\top\bm{r}_{\rm{pl/sc}}(t)^{}\left(c^2 - \bm{v}_{\rm{pl}}(t)^\top\bm{v}_{\rm{pl}}(t)^{}\right) + \left(\bm{r}_{\rm{pl/sc}}(t)^\top \bm{v}_{\rm{pl}}(t)^{} \right)^2}\right) 
\end{aligned}
\end{equation}

Eq.\ \eqref{eq:twosolutions} shows that two solutions are possible, given the geometry between the planet and the spacecraft. It is important to understand which solution is the correct one to uniquely solve for $\Delta t$. By defining
 $\bm{\beta}_{\rm{pl}}(t) = \dfrac{\bm{v}_{\rm{pl}}(t)}{c}$ and the planet flight path angle $\epsilon$, the approximated solution for the light-time correction is:

\begin{equation}
\begin{aligned}
    \Delta t %
    &= \frac{ - c \,\left| \left| \bm{r}_{\rm{pl/sc}}(t) \right| \right|\left| \left| \bm{\beta}_{\rm{pl}}(t) \right| \right| \cos\epsilon \pm c \,\left| \left| \bm{r}_{\rm{pl/sc}}(t) \right| \right|  \sqrt{\left| \left| \bm{\beta}_{\rm{pl}}(t) \right| \right|^2 \left(\cos^2\epsilon - 1 \right)  + 1 }}{c^2\left(1 - \left| \left| \bm{\beta}_{\rm{pl}}(t) \right| \right|^2 \right)}
\end{aligned}
\end{equation}

Recall that the correct solution is the one providing $\Delta t \geq 0$ as the light departs from the planet before arriving at the spacecraft camera. As a consequence, the solution with the plus sign is the one providing the correct time delay. Thus:
\begin{equation}\label{eq:deltaT}
    \Delta t = \frac{\left| \left| \bm{r}_{\rm{pl/sc}}(t) \right| \right|}{c\left(1 - \left| \left| \bm{\beta}_{\rm{pl}}(t) \right| \right|^2 \right)} \left( - \left| \left| \bm{\beta}_{\rm{pl}}(t) \right| \right|\cos\epsilon + \sqrt{\left| \left| \bm{\beta}_{\rm{pl}}(t) \right| \right|\left(\cos^2\epsilon -1\right) +1} \right)
\end{equation}
It is worthy noting $1 - \left| \left| \bm{\beta}_{\rm{pl}}(t) \right| \right|^2 \geq 0 $ $\forall \bm{\beta}_{\rm{pl}}(t)$ and $\left| \left| \bm{\beta}_{\rm{pl}}(t) \right| \right|^2 \left(\cos^2\epsilon - 1 \right)  + 1 \geq 0$ $\forall \bm{\beta}_{\rm{pl}}(t)$ and $\forall \epsilon$, as $ \left| \left| \bm{\beta}_{\rm{pl}}(t) \right| \right| \leq 1$. Note that $\cos\epsilon\geq 0$ $\forall\epsilon$ by flight path angle definition. 
Eq.\ \eqref{eq:deltaT} provides an analytical solution at first order for the light-time delay which can be exploited to include light-time correction in the filter update.
\subsubsection{Definition of the measurement model equation}
Once $\Delta t$ is computed, the planet LoS can be expressed as the unit vector for the spacecraft position at time $t$ to the planet position at time $\tau$. Thus:
\begin{equation}
    \bm{l}_{\rm{pl/sc}} = \frac{\left(\bm{r}_{\rm{pl}}(t-\Delta t) -\bm{r}(t)\right)^\top \left(\bm{r}_{\rm{pl}}(t-\Delta t) -\bm{r}(t)\right)}{\left|\left|\left(\bm{r}_{\rm{pl}}(t-\Delta t) -\bm{r}(t)\right)^\top \left(\bm{r}_{\rm{pl}}(t-\Delta t) -\bm{r}(t)\right)\right|\right|}
\end{equation}

This unit vector is warped by relativistic light aberration as the spacecraft is not fixed with respect to the inertial reference frame. At first order, this effect can be expressed as follows \cite{shuster2003stellar}:
\begin{equation}\label{eq:LOSaberr}
    {}^{\mathcal{}}\bm{l}_{\rm{pl/sc}}^{\rm{aberr}} = {}^{\mathcal{}}\bm{l}_{\rm{pl/sc}} + {}^{\mathcal{}}\bm{l}_{\rm{pl/sc}}\times\left({}^{\mathcal{}}\bm{\beta}_{\rm{sc}}\times {}^{\mathcal{}}\bm{l}_{\rm{pl/sc}}\right)
\end{equation}
where ${}^{\mathcal{}}\bm{\beta}_{\rm{sc}} = \dfrac{ {}^{\mathcal{}}\bm{v}}{c}$. Note that higher orders are not detectable from the image processing pipeline as they are orders of magnitude below 15 arcsec \cite{christian2019starnav}, which is attitude determination performance. 

Finally, the warped line of sight is projected in the camera:
\begin{equation}\label{eq:projection}
    {}_h^{\mathbb{C}} \bm{r}_{\rm{pl}} = \bm{K}_{\rm{cam}}\,\bm A_{\rm{corr}} {}^{\mathcal{}}\bm{l}_{\rm{pl/sc}}^{\rm{aberr}}
\end{equation}
\begin{equation}
    {}^{\mathbb{C}} \bm{r}_{\rm{pl}} = \frac{1}{{}_h^{\mathbb{C}} \bm{r}_{\rm{pl}, (3)}}\begin{pmatrix}{}_h^{\mathbb{C}} \bm{r}_{\rm{pl}, (1)}\\{}_h^{\mathbb{C}} \bm{r}_{\rm{pl}, (2)}\end{pmatrix}
\end{equation}
where ${}_h^{\mathbb{C}} \bm{r}_{\rm{pl}}$ is the projection of the planet line of sight in the image plane in homogeneous coordinates, ${}^{\mathbb{C}} \bm{r}_{\rm{pl}}$ is the same vector but in non-homogeneous coordinates, ${}_h^{\mathbb{C}} \bm{r}_{\rm{pl}, (i)}$ is the $i$-esimal coordinate of vector ${}_h^{\mathbb{C}} \bm{r}_{\rm{pl}}$, $\bm{K}_{\rm{cam}}$ is the camera intrinsic matrix, and $\bm A_{\rm{corr}}$ is the rotation matrix from the inertial reference frame $\mathcal{N}$ to the camera reference frame $\mathcal{C}$ corrected for the stars light aberration.

\subsubsection{Definition of the Jacobian of the measurement model}
As the measurement model is analytic, its Jacobian can be easily computed by variational analysis. Thus:
\begin{equation}\label{eq:proj}
    \delta \, {}^{\mathbb{C}} \bm{r}_{\rm{pl}} = \begin{bmatrix}
         \dfrac{1}{{}_h^{\mathbb{C}} \bm{r}_{\rm{pl}, (3)}} & 0 &  -\dfrac{{}_h^{\mathbb{C}} \bm{r}_{\rm{pl}, (1)}}{{}_h^{\mathbb{C}} \bm{r}_{\rm{pl}, (3)}^{2}}\\
         0& \dfrac{1}{{}_h^{\mathbb{C}} \bm{r}_{\rm{pl}, (3)}} & -\dfrac{{}_h^{\mathbb{C}} \bm{r}_{\rm{pl}, (2)}}{{}_h^{\mathbb{C}} \bm{r}_{\rm{pl}, (3)}^{2}}
    \end{bmatrix} \, \delta {}_h^{\mathbb{C}} \bm{r}_{\rm{pl}}
\end{equation}
\begin{equation}\label{eq:projHom}
   \delta {}_h^{\mathbb{C}} \bm{r}_{\rm{pl}} =  \bm{K}_{\rm{cam}}\bm A_{\rm{corr}}\left( \delta {}^{\mathcal{}}\bm{l}_{\rm{pl/sc}}^{\rm{aberr}} + 2 \left[ {}^{\mathcal{}}\bm{l}_{\rm{pl/sc}}^{\rm{aberr}}\right]^{\wedge} \delta \bm{q}_{\mathcal{C}/\mathcal{N}}^{(v)} \right)
\end{equation}
where $\bm{q}_{\mathcal{C}/\mathcal{N}}^{(v)}$ is the vectorial part of the quaternion representing the rotation from the inertial reference frame $\mathcal{N}$ to the camera reference frame $\mathcal{C}$. 

The variation of the aberrated line of sight ${}^{\mathcal{}}\bm{l}_{\rm{pl/sc}}^{\rm{aberr}}$ is computed by exploiting the triple vector product identity  $\mathbf {a} \times (\mathbf {b} \times \mathbf {c} )=(\mathbf {a} \cdot \mathbf {c} )\mathbf {b} -(\mathbf {a} \cdot \mathbf {b} )\mathbf {c}$:
\begin{equation}\label{eq:deltaLOSaberr}
\begin{aligned}
    \delta {}^{\mathcal{}}\bm{l}_{\rm{pl/sc}}^{\rm{aberr}} = &\left(\bm I_{3x3} + 2\, {}^{\mathcal{}}\bm{\beta}_{\rm{sc}} {}^{\mathcal{}}\bm{l}_{\rm{pl/sc}}^\top - {}^{\mathcal{}}\bm{l}_{\rm{pl/sc}}^\top  {}^{\mathcal{}}\bm{\beta}_{\rm{sc}}- {}^{\mathcal{}}\bm{l}_{\rm{pl/sc}}{}^{\mathcal{}}\bm{\beta}_{\rm{sc}}^\top \right) \delta {}^{\mathcal{}}\bm{l}_{\rm{pl/sc}} +\\&+ \left({}^{\mathcal{}}\bm{l}_{\rm{pl/sc}}^\top  {}^{\mathcal{}}\bm{l}_{\rm{pl/sc}} - {}^{\mathcal{}}\bm{l}_{\rm{pl/sc}}{}^{\mathcal{}}\bm{l}_{\rm{pl/sc}}^\top  \right) \delta {}^{\mathcal{}}\bm{\beta}_{\rm{sc}}
\end{aligned}
\end{equation}
where $\delta {}^{\mathcal{}}\bm{\beta}_{\rm{sc}} = \dfrac{ \delta {}^{\mathcal{}}\bm{v}}{c}$ and $ \delta {}^{\mathcal{}}\bm{l}_{\rm{pl/sc}}$ is:
\begin{multline}\label{eq:deltaLOS}
    \delta {}^{\mathcal{}}\bm{l}_{\rm{pl/sc}} = \left(  \frac{\bm{I}_{3x3}}{\left|\left|\left(\bm{r}_{\rm{pl}}(t-\Delta t) -\bm{r}(t)\right)^\top \left(\bm{r}_{\rm{pl}}(t-\Delta t) -\bm{r}(t)\right)\right|\right|} +\right.\\ \left.- \frac{\left(\bm{r}_{\rm{pl}}(t-\Delta t) -\bm{r}(t)\right) \left(\bm{r}_{\rm{pl}}(t-\Delta t) -\bm{r}(t)\right)^\top}{\left|\left|\left(\bm{r}_{\rm{pl}}(t-\Delta t) -\bm{r}(t)\right)^\top \left(\bm{r}_{\rm{pl}}(t-\Delta t) -\bm{r}(t)\right)\right|\right|^3}\right) \left(\delta\bm{r}_{\rm{pl}}(t-\Delta t) -\delta\bm{r}(t) \right) 
\end{multline}
The variation of the observed LoS from the camera can be gathered as follows:
\begin{equation}\label{eq:deltaPlanetPosLT}
    \delta\bm{r}_{\rm{pl}}(t-\Delta t) = \delta\bm{r}_{\rm{pl}}(t) - \Delta t\,\delta\bm{v}_{\rm{pl}}(t) - \bm{v}_{\rm{pl}}(t) \delta\Delta t
\end{equation}
where 
\begin{multline}\label{eq:deltaDeltaT}
    \delta\Delta t = \frac{1}{c^2 - \bm{v}_{\rm{pl}}(t)^\top\bm{v}_{\rm{pl}}(t)} \Biggl( \Biggl( \frac{ \left(c^2 - \bm{v}_{\rm{pl}}(t)^\top\bm{v}_{\rm{pl}}(t)\right)\bm{r}_{\rm{pl/sc}}(t)^\top + \bm{r}_{\rm{pl/sc}}(t)^\top \bm{v}_{\rm{pl}}(t)^{} \bm{v}_{\rm{pl}}(t)^\top }{\sqrt{\bm{r}_{\rm{pl/sc}}(t)^\top\bm{r}_{\rm{pl/sc}}(t)^{}\left(c^2 - \bm{v}_{\rm{pl}}(t)^\top\bm{v}_{\rm{pl}}(t)^{}\right) + \left(\bm{r}_{\rm{pl/sc}}(t)^\top \bm{v}_{\rm{pl}}(t)^{} \right)^2}} \\
    - \bm{v}_{\rm{pl}}^\top(t)\Biggr) \left(\delta \bm{r}_{\rm{pl}}^{}(t) - \delta\bm{r}(t)\right) + \Biggl( \frac{ 
    \bm{r}_{\rm{pl/sc}}(t)^\top \bm{v}_{\rm{pl}}(t)^{} \bm{r}_{\rm{pl/sc}}(t)^\top - \bm{r}_{\rm{pl/sc}}(t)^\top\bm{r}_{\rm{pl/sc}}(t) \bm{v}_{\rm{pl}}(t)^\top}{\sqrt{\bm{r}_{\rm{pl/sc}}(t)^\top\bm{r}_{\rm{pl/sc}}(t)^{}\left(c^2 - \bm{v}_{\rm{pl}}(t)^\top\bm{v}_{\rm{pl}}(t)^{}\right) + \left(\bm{r}_{\rm{pl/sc}}(t)^\top \bm{v}_{\rm{pl}}(t)^{} \right)^2}} +\\
    + \frac{2 \bm{v}_{\rm{pl}}(t)^\top }{c^2 - \bm{v}_{\rm{pl}}(t)^\top\bm{v}_{\rm{pl}}(t)}
    -\bm{r}_{\rm{pl/sc}}(t)^\top\Biggr) \delta \bm{v}_{\rm{pl}}(t) \Biggr)
\end{multline}
Finally, by combining Eqs.\ \eqref{eq:proj}--\eqref{eq:deltaDeltaT}, the linear mapping between the variation of ${}^{\mathbb{C}} \bm{r}_{\rm{pl}}$ and the variation of $\bm{r}_{\rm{pl}}$, $\bm{v}_{\rm{pl}}$, $\bm{r}$, and $\bm{q}_{\mathcal{C}/\mathcal{N}}^{(v)}$ can be established:
\begin{equation}\label{eq:jacMeas}
    \delta {}^{\mathbb{C}} \bm{r}_{\rm{pl}} = \bm{\Pi} \begin{pmatrix}
        \delta \bm{r}(t) \\ \delta \bm{v}(t) \\ \delta \bm{q}_{\mathcal{C}/\mathcal{N}}^{(v)} \\ \delta \bm{r}_{\rm{pl}}(t) \\ \delta \bm{v}_{\rm{pl}}(t)
    \end{pmatrix}
\end{equation}
Eq.\ \eqref{eq:jacMeas} provides the linear mapping between the projection of the planet in the image plane as a function of the spacecraft state and of the planet motion in the solar system. In an operational scenario, the variation of the spacecraft state comes from the orbital and attitude filter performance, whereas the variation of the planet motion arises from the knowledge of the planet ephemeris or from the approximation errors induced by an onboard implementation of the planet motion. The Jacobian matrix of the measurement model exploited in the navigation filter takes into account the first six columns of $\bm \Pi$, being only the probe position and velocity contained in the state vector $\bm x$.

The advantages of the proposed measurement model are threefold. Firstly, the proposed model avoids any operation on the raw measurement extracted from the deep-space images to correct the light-aberration effect. %
Secondly, the expression found in Eq.\ \eqref{eq:deltaT} is analytic and provides a first-order approximation for the light-time delay which can be used to find efficiently on board $\Delta t$ without any optimization, at the contrary of \cite{andreis2022onboard}. Thirdly, the proposed equation models the light effects as a function of the spacecraft and planet states, implying that their uncertainty can be taken into account during filter updates.

In the following, two approaches to correct the light-aberration effect are analyzed. Case 1 exploits the complete measurement model developed in this section. Whereas, in case 2 only the light-time correction is included in the measurement model, as the relativistic light aberration is compensated when performing attitude determination on all the bright objects in the image (see Sec.\ \ref{sec:planetObs}). This implies that Eqs.\ \eqref{eq:LOSaberr} and \eqref{eq:deltaLOSaberr} are not used and that Eqs.\ \eqref{eq:projection} and \eqref{eq:projHom} simply become:
\begin{equation}
    {}_h^{\mathbb{C}} \bm{r}_{\rm{pl}} = \bm{K}\bm A_{\rm{corr}}{}^{\mathcal{}}\bm{l}_{\rm{pl/sc}}
\end{equation}
\begin{equation}
   \delta {}_h^{\mathbb{C}} \bm{r}_{\rm{pl}} =  \bm{K}\bm{A}_{\rm{corr}}\left( \delta {}^{\mathcal{}}\bm{l}_{\rm{pl/sc}} + 2 \left[ {}^{\mathcal{}}\bm{l}_{\rm{pl/sc}}\right]^{\wedge} \delta \bm{q}_{\mathcal{C}/\mathcal{N}}^{(v)} \right)
\end{equation}
\subsection{Selected Filtering Strategy for the Vision-Based Navigation Algorithm}
A non-dimensionalized EKF is selected as the most appropriate filtering approach for the development of a VBN algorithm for CubeSat applications. The selection has been performed in \cite{andreis2022onboard}, where the behavior of five different EKFs has been analyzed in terms of estimator numerical stability and computational performance. Indeed, it is important to remark that the autonomous VBN algorithm has to be deployed on a miniaturized processor characterized by limited computation capabilities comparable to the one of a Raspberry Pi. The implemented scheme is reported in \cref{tab:filter_scheme}, where all the terms are already non-dimensionalized following the approach discussed in~\cite{andreis2022onboard}. 
\begin{table}[H]
     \centering
      \caption{Filtering Strategy}
\begin{tabular}{p{1.2in}p{2.1in}p{1.4in}p{0.4in}}
\toprule \toprule
System State Space &\multicolumn{2}{c}{$ \dot{\bm{x}} = \bm f(\bm x(t),t) + \bm w$} & \\
& \multicolumn{2}{c}{$\bm y_k = \bm h(\bm x_k) + \bm \nu_k$} &  \\
& \multicolumn{2}{c}{ $\dot{\bm {P}} = \bm F \bm P + \bm P \bm F^\top + \bm Q$}&  \\
\midrule
Propagation Block & $\bm x_{p_{k}} = \bm x_{c_{k-1}} + \int_{t_{k-1}}^{t_{k}}\bm f(\bm x(t),t)\textrm{d}t$ & $\bm x_{c_0} = E[\bm x_0]$&  \\
& $\bm{P}_{p_k}=  \bm{P}_{c_{k-1}}+ \int_{t_{k-1}}^{t_{k}}{ \dot{\bm P} \textrm{d}t}$ & $\bm P_{c_0} = E[\bm x_0 \bm x_0^\top]$ &  \\
\midrule
Correction Block &\multicolumn{2}{c}{$ \bm K_k = \bm P_{p_k} \bm H_k^\top( \bm H_k \bm P_{p_k} \bm H_k^\top + \bm R_k)^{-1} $}&\\
& \multicolumn{2}{c}{$ \bm x_{c_k} = \bm x_{p_k} + \bm K_k[\bm y_k- \bm h(\bm x_{p_k})] $}&\\
& \multicolumn{2}{c}{$\bm P_{c_k} = (\bm I - \bm K_k \bm H_k)\bm P_{p_k}(\bm I - \bm K_k \bm H_k)^\top + \bm K_k \bm R_k \bm K_k^\top$}&\\
\bottomrule 
\bottomrule
\end{tabular}    
     \label{tab:filter_scheme}
\end{table}
Here, $\bm x_{p_k}$ is the predicted state vector with error covariance matrix $\bm P_{p_k}$ at epoch $k$, $\bm K_k$ the Kalman gain, $\bm x_{c_k}$ the corrected state vector with error covariance matrix $\bm P_{c_k}$, $\bm F$ the Jacobian of the dynamics model equation, $\bm h$ the measurement model equation with Jacobian $\bm H_k$, $\bm \nu_k$ the measurement white noise, and $\bm y_k$ the external measurement vector.

Moreover, two additional procedures are implemented in the navigation filter to face the errors of the IP algorithm: 1) When observations are not acquired due to an IP failure, the state vector and its error covariance matrix are simply propagated until the next step; 2) an innovation-based outlier detection method is applied to reject false positives \cite{liu2004line}. In particular, when the absolute value of the innovation term ($|| \bm r_{\rm{pl}_k}- \bm h(\bm x_{p_k})|| $) is greater than $k\sqrt{\bm M_{ii}}$ with 
$\bm M = \bm H_k \bm P_{p_k} \bm H_k^\top + \bm R_k$ and $k = 3$, the innovation term is set to zero, and the filter correction step is not performed. Indeed, it is preferred to keep an old but good prediction so as not to worsen the estimation.
\section{Results}
\label{sec: results}
\subsection{Image Processing Performance}
\label{sec:image processing results}
\begin{figure}[h!]
\centering
\begin{subfigure}{0.5\textwidth}
\centering
 \includegraphics[width = \textwidth]{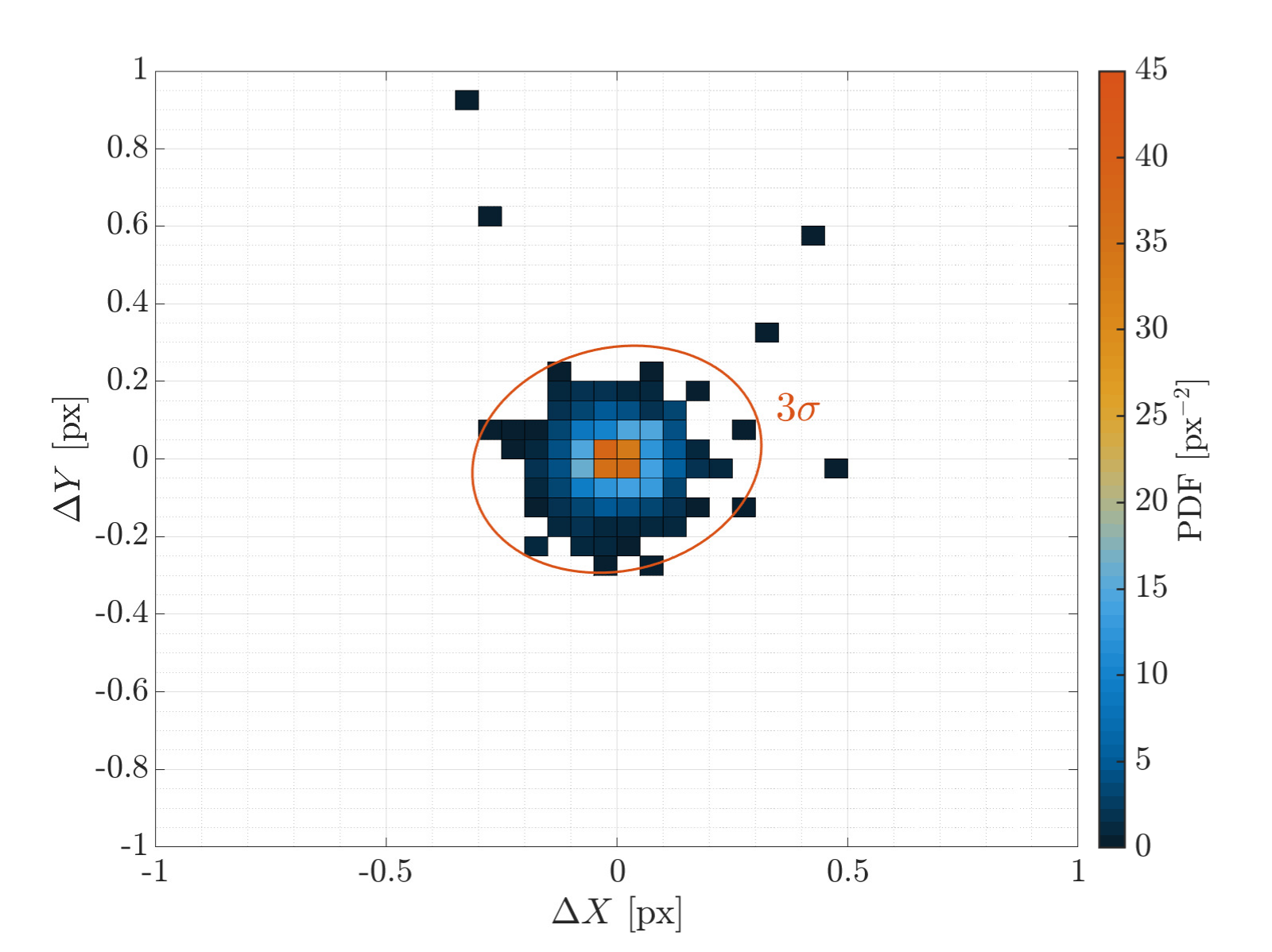}
 \caption{$\sigma_{r} = 10^4$ km}
 \label{fig:sigmar_1e4}
 \end{subfigure}
\begin{subfigure}{0.49\textwidth}
\centering
 \includegraphics[width = \textwidth]{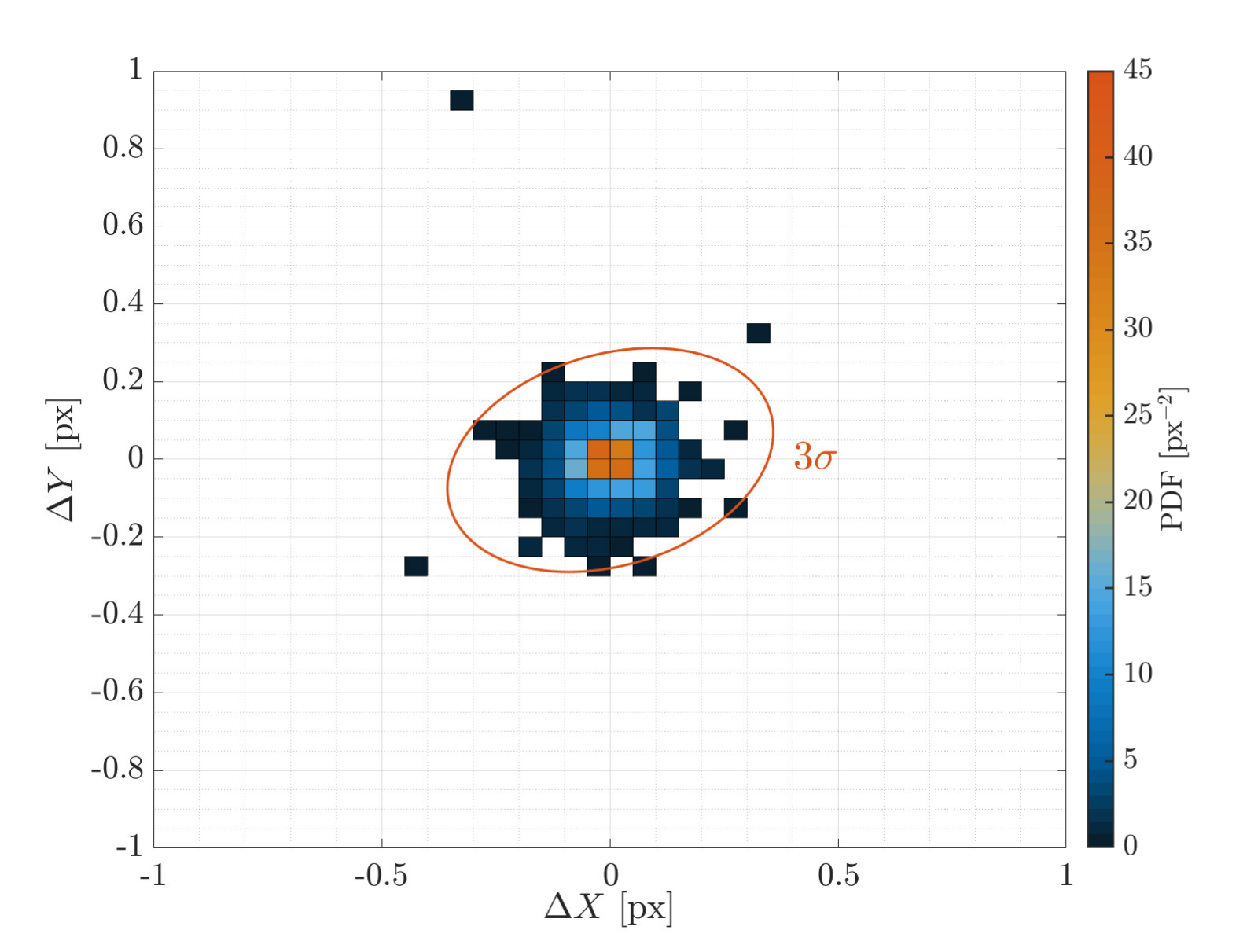}
 \caption{$\sigma_{r} = 10^5$ km}
 \label{fig:sigmar_1e5}
 \end{subfigure}
\begin{subfigure}{0.49\textwidth}
\centering
 \includegraphics[width = \textwidth]{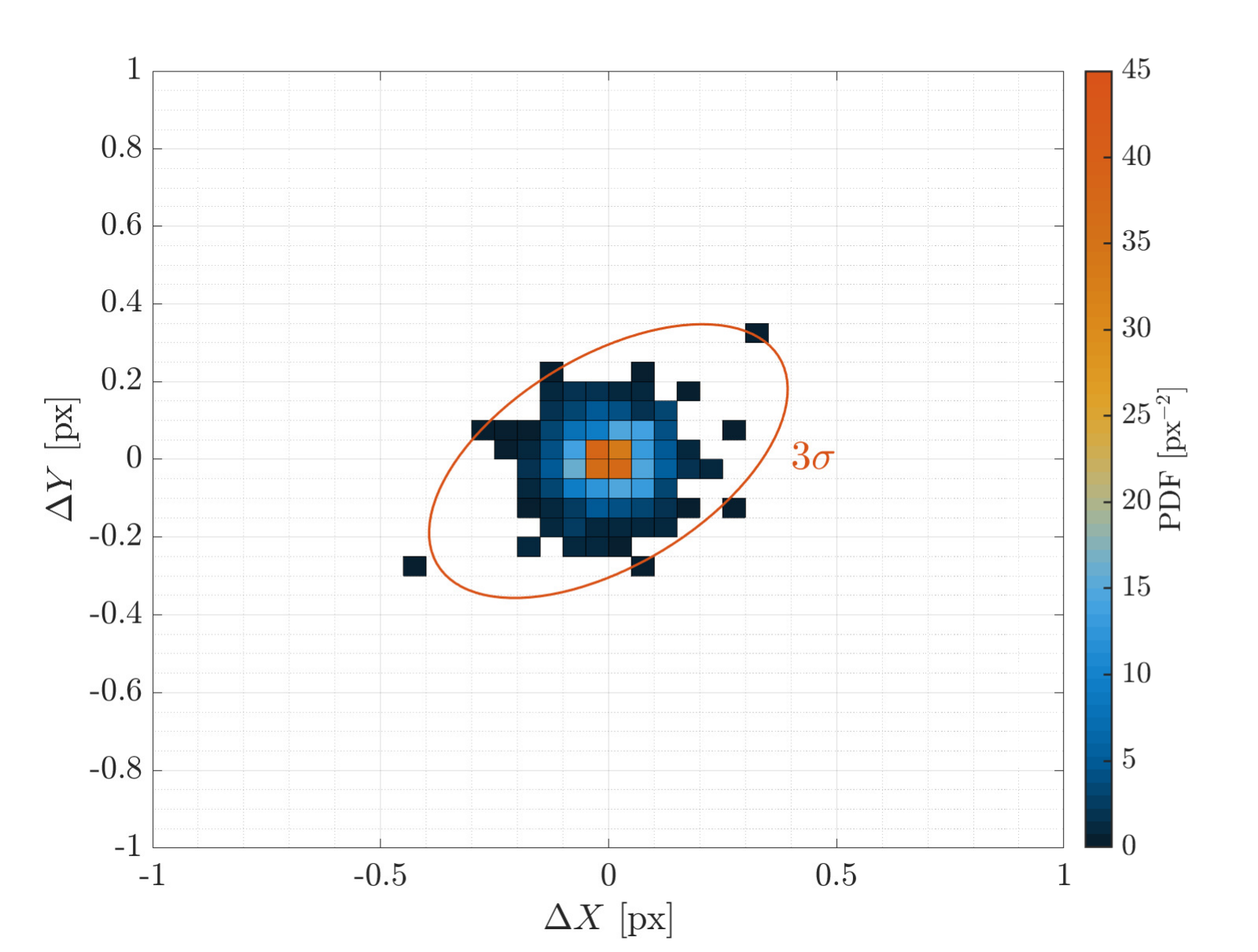}
 \caption{$\sigma_{r} = 10^6$ km}
 \label{fig:sigmar_1e6}
 \end{subfigure}
\begin{subfigure}{0.49\textwidth}
\centering
 \includegraphics[width = \textwidth]{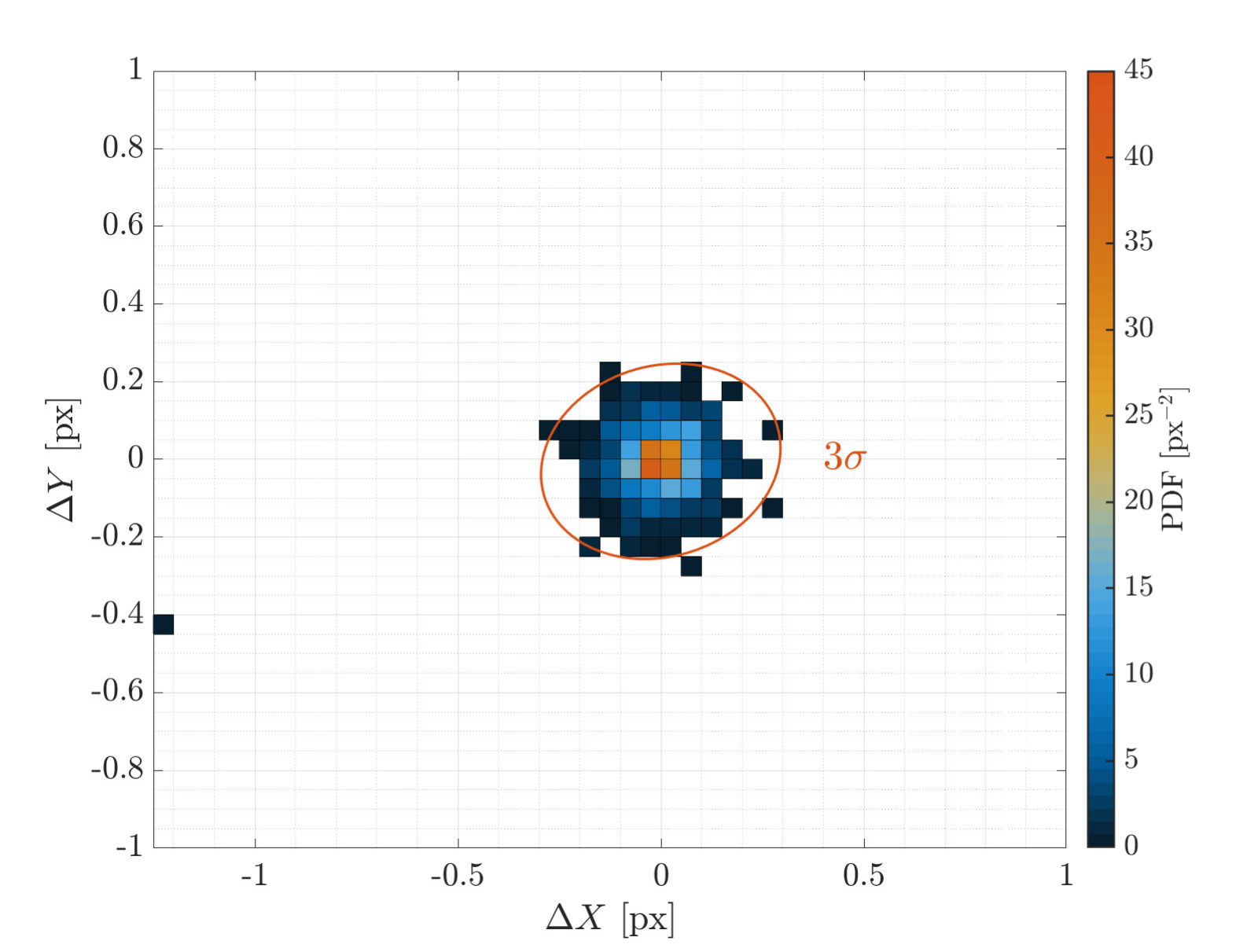}
 \caption{$\sigma_{r} = 10^7$ km}
 \label{fig:sigmar_1e7}
 \end{subfigure}
\caption{Probability Density Function of the planet position projection errors with 3$\sigma$ bounds.}
\label{fig:gauss_distr}
\end{figure}

To validate the IP algorithm performance before adopting it inside the VBN filter, four Monte Carlo campaigns are carried out where the initial uncertainty of the probe position $\sigma_{\bm r}$ is set to $10^4$, $10^5$, $10^6$, and $10^7$ km, respectively. In each campaign, the extraction of the beacon position projection is run for 1031 scenarios, wherein at least one planet is present, out of the 3000 analyzed. In each scenario, the position of the spacecraft is selected by randomly sampling a Gaussian distribution with $\sigma_x=\sigma_y=3 \rm{AU}$ and $\sigma_z = 0.07  \rm{AU}$ and centered in the origin of $\mathcal{N}$. The $z$-component of the probe position is chosen in a narrower interval as the spacecraft is supposed to lie close to the ecliptic plane. Similarly, the orientation of the probe is assigned by randomly sampling a normal distribution $\alpha$, $\delta$, and $\phi$ in the $3\sigma$ intervals $[0,2\pi]$, $[-0.6,0.6]$, and $[0,2\pi]$, respectively. The declination $\delta$ is chosen in a narrower interval as planets are distributed close to the ecliptic plane.  
For what concerns the planet detection step, $\sigma_{\bm q_v}$ is set equal to 20 arcsec as a result of a statistical analysis conducted on the error obtained in the attitude determination. The planet position uncertainty $\sigma_{\bm r_{\rm{pl}}}$ is assumed equal to zero because of the high accuracy with which the planets ephemeris are known. %
\newline
The performance indexes adopted for the discussion are the angular error for the attitude determination and the beacon projection error for the planet detection step.
The Probability Density Function (PDF) along with the 3$\sigma$ ellipsoid of the best-fit Gaussian distribution for $\sigma_{\bm r} = 10^4$ km, $\sigma_{\bm r} = 10^5$ km, $\sigma_{\bm r} = 10^6$ km, and $\sigma_{\bm r} = 10^7$ km are shown in Figs.\ \ref{fig:sigmar_1e4}, \ref{fig:sigmar_1e5}, \ref{fig:sigmar_1e6}, and \ref{fig:sigmar_1e7}, respectively.
Note that the planet position projection is detected with a sub-pixel $3\sigma$ accuracy for all the values of $\sigma_{\bm r}$. In other terms, the error on the estimated planet position projection is not dependent on the probe position uncertainty but only on the attitude determination and centroids computation errors.
When the probe position uncertainty increases, the scenarios where the beacon projection error is over 0.3 px seem to be filtered out. Indeed, in these cases, the IP algorithm may select a different wrong spike as the expected position projection becomes far from the real one. As a result, the error norm becomes greater than the threshold value set for the assessment of the IP performance. This case is considered to be a failure of the IP pipeline and it is not represented in the PDF. Note that this condition is only applied during the IP testing, therefore, it is not present when the filter is run.
The $3\sigma$ error ellipses in Fig.\ \ref{fig:gauss_distr} are obtained from the mean and covariance values reported in Table \ref{tab:covariance}. 
\begin{table}[h!]
    \centering
     \caption{Mean and covariance of the planet position projection errors for the four values of $\sigma_{\bm r}$}
     \begin{tabular}{c|cccc}
          \toprule \toprule
$\sigma_{\bm r}$ [km] & $10^4$ & $10^5$ & $10^6$ & $10^7$\\
\hline
& & & & \\
$\bm P_{\rm{err}}$ $[\rm{px}^2]$& 
 $\begin{bmatrix}
   0.008	& 0.001\\
   0.001	& 0.007\\
\end{bmatrix} $&  
 $\begin{bmatrix}
    0.011 &	0.002\\
    0.002 &	0.007 \\
\end{bmatrix}$  & 
 $\begin{bmatrix}
    0.013 &	0.006\\ 
    0.006 &	0.011 \\
\end{bmatrix} $ & 
 $\begin{bmatrix}
    0.007 &	0.001\\
    0.001 &	0.005 \\
\end{bmatrix} $\\

$\bm \mu_{\rm{err}}$ [px] & [0.0014;-0.0003] & [0.0001;-0.0034] & [-0.0005;	-0.0045] & [-0.0005;-0.0041] \\

$\text{det}(\bm P) $[\rm{px}$^4$] & 5.9e-05 &  7.05e-05 & 9.94e-05 &  7.05e-05 \\
 \bottomrule \bottomrule
    \end{tabular}
     \label{tab:covariance}
\end{table}
The four covariance matrices are characterized by a similar determinant, which is proportional to the area of the ellipse, implying that the precision in the planet determination is not dependent on the uncertainty of the spacecraft position. This feature is one of the advantages of the proposed pipeline for planet detection in deep-space images.
The results of the IP robustness and attitude determination error are shown in Table\ \ref{tab: 0spike} for the 962 cases in which a correct or wrong attitude solution is found.

\begin{table}[h!]
    \centering
     \caption{Algorithm Performance}
     \begin{tabular}{c|ccc}
          \toprule \toprule
\makecell{ $\sigma_{r} $ \\ $\text{[km]}$ } & \makecell{$\sigma_{\rm{ErrRot}}$ \\$\text{[arcsec]}$} & \makecell{\% Wrong Beacon \\ Detection \\ (of 962  cases)} & \makecell{\% Wrong Beacon Detection with \\ Right Attitude Determination \\ (of 962  cases)} \\
\hline
$10^4$ &  14.77  & 4.57 (44 cases) & 0.42 (4 cases)\\
$10^5$ & 15.18 & 4.68 (45 cases) & 0.52 (5 cases) \\
$10^6$ & 15.21 &  7.69 (74 cases) & 3.33 (32 cases)\\
$10^7$ & 15.48 & 29.83 (287 cases) & 25.47 (245 cases)\\
 \bottomrule \bottomrule
    \end{tabular}
     \label{tab: 0spike}
\end{table}

\begin{figure}[h!]
\begin{subfigure}{0.48\textwidth}
\includegraphics[width =\textwidth]{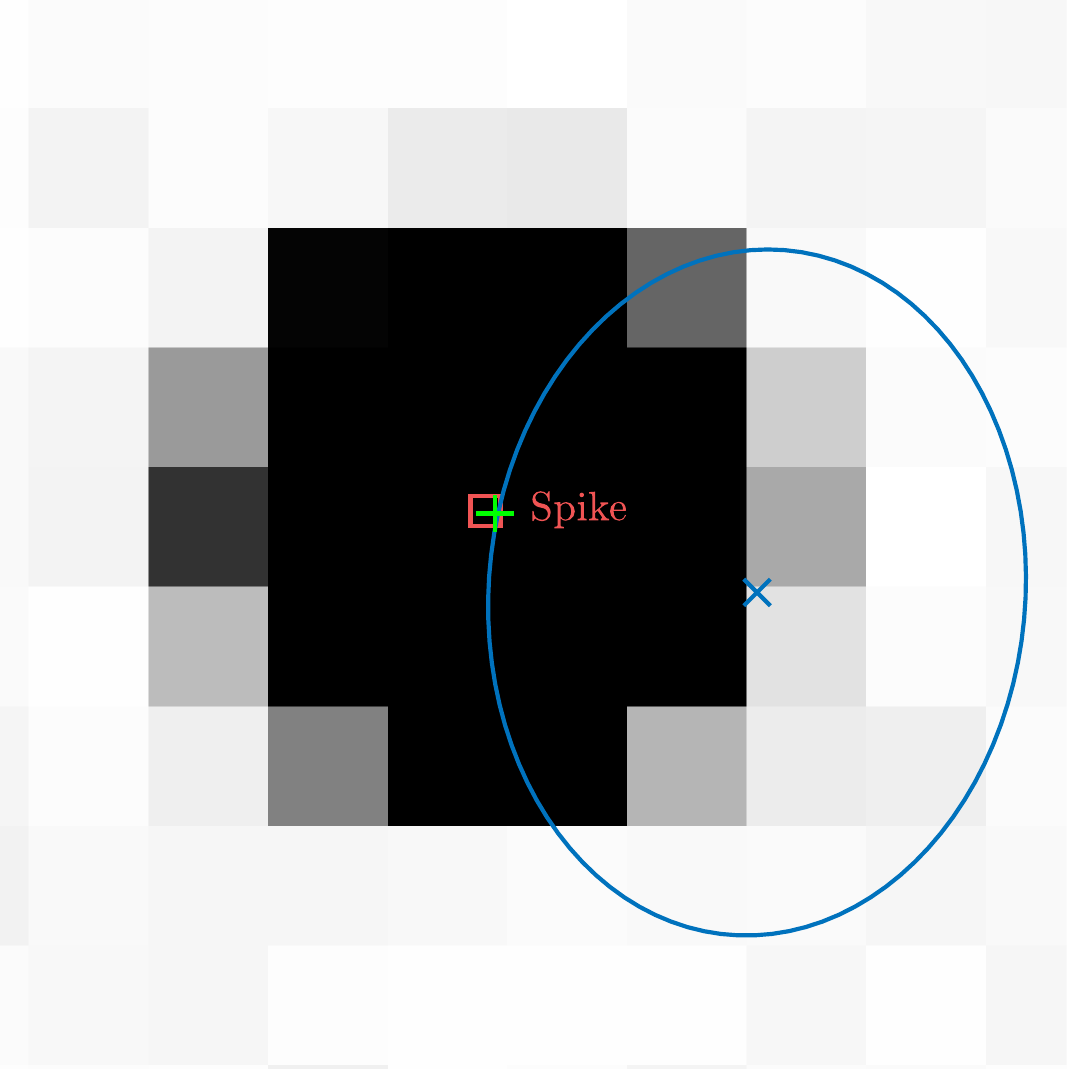}
\caption{ The spike related to the planet projection falls outside the 3$\sigma$ ellipse.}
\label{fig:Failure Case III.B}
\end{subfigure}
\begin{subfigure}{0.48\textwidth}
\includegraphics[width =\textwidth]{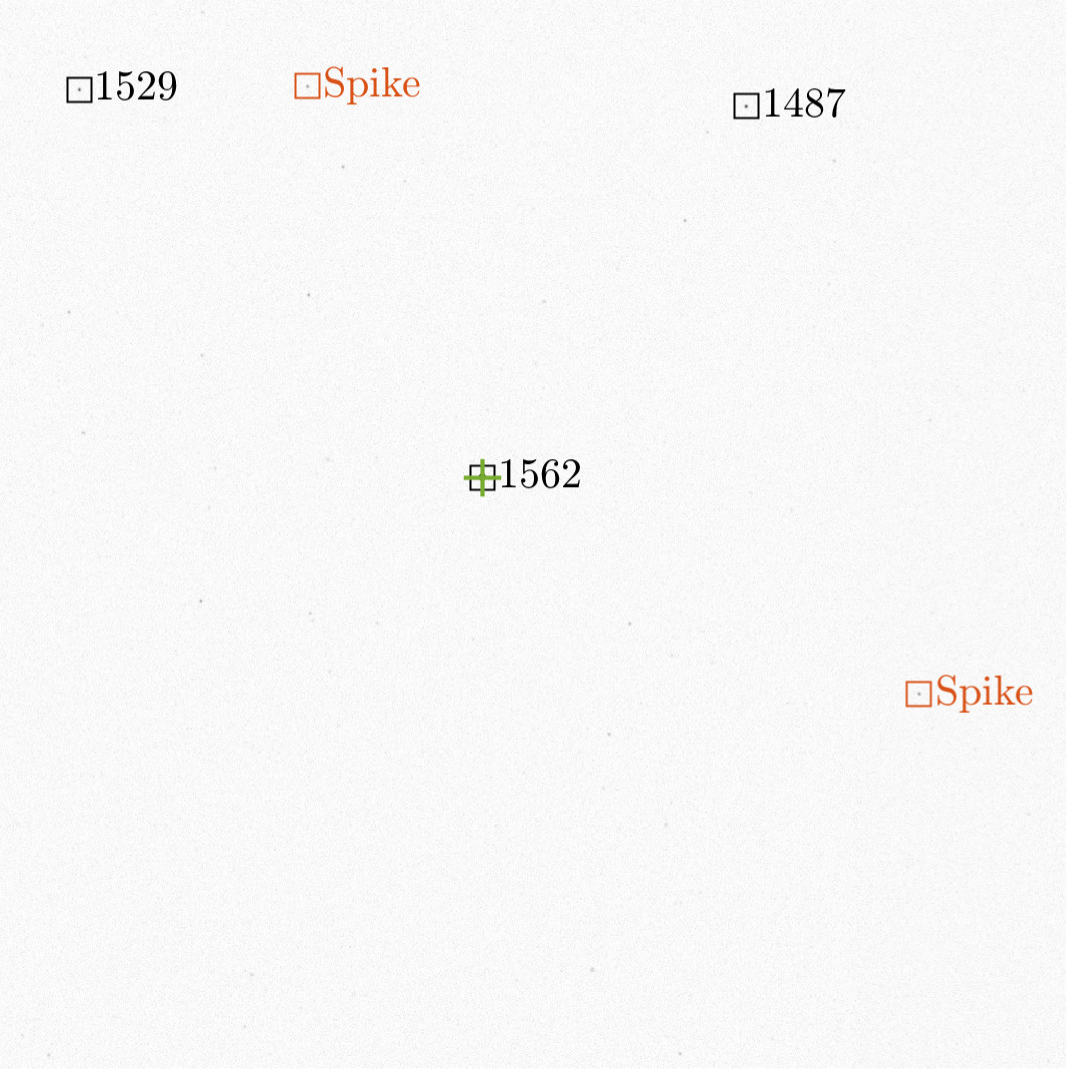}
\caption{The attitude is wrongly determined. The planet is mistaken for star 1562. }
\label{fig:Failure Case III.C}
\end{subfigure}
\begin{subfigure}{0.48\textwidth}
\centering
 \includegraphics[width = \textwidth]{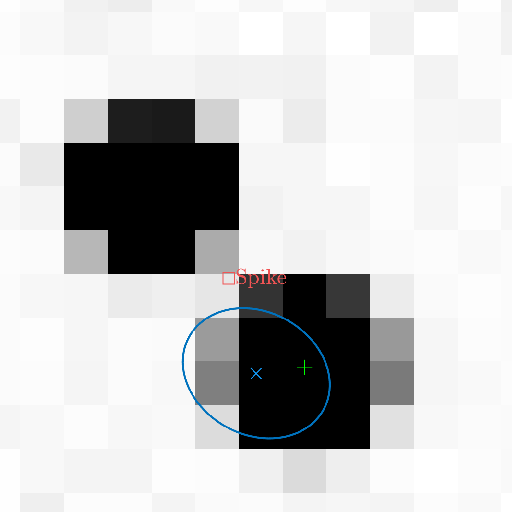}
 \caption{The centroiding algorithm resolves only one centroid, instead of two.}
 \label{fig:Failure Case III.E}
 \end{subfigure}
\hspace{0.5cm}
\begin{subfigure}{0.48\textwidth}
\centering
 \includegraphics[width = \textwidth]{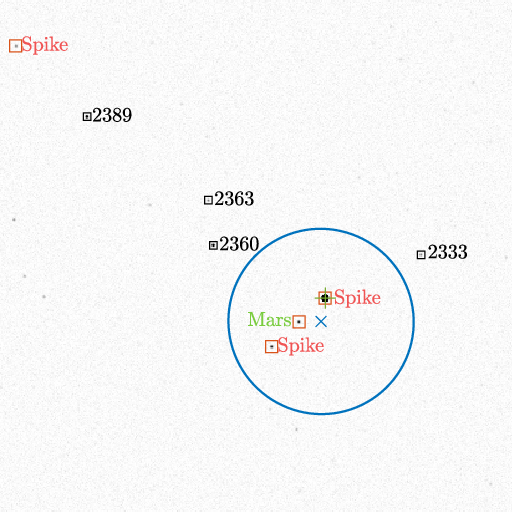}
 \caption{Planet wrongly identified. The planet is associated with a wrong spike.}
\label{fig:second scenario}
 \end{subfigure}
\caption{Scenarios in which the IP pipeline fails in the planet detection.}
\label{fig:IP fail}
\end{figure}

The percentage of off-nominal scenarios during planet identification greatly depends on the probe position uncertainty. Indeed, when $\sigma_{\bm r}$ increases, the expected planet position projection is further from the real position projection, and its uncertainty ellipse is bigger, which leads to a higher probability of planet misidentification. Moreover, the percentage of off-nominal scenarios in planet detection also depends strictly on the success of the attitude determination. Indeed, when attitude determination provides the wrong solution, planet detection fails consequentially. In Table~\ref{tab: 0spike}, the third column represents the total number of cases of wrong planet identification when the attitude determination converges to a right or wrong solution. Instead, the last column represents the number of cases of wrong planet identification when the attitude determination converges to the correct solution. The failure percentage of the beacon detection procedure when the probe attitude is correctly determined is lower than 1\% with a probe position uncertainty up to $10^5$ km. 
\newline
To sum up, the accuracy of the proposed method is independent of the probe position uncertainty, and it relies only on the centroid errors. Whereas, the robustness of the IP depends on the attitude determination performance and on the probe position uncertainty. If a more robust star identification algorithm is adopted, such as the pyramid one \cite{mortari2004pyramid}, or an attitude filter is implemented, the total percentage of failure in the beacon detection becomes remarkably lower.
For completeness' sake, Fig.\ \ref{fig:IP fail} shows some scenarios that have been found during the assessment of the IP performance where the procedure fails in planet detection. In particular, in Figs.\ \ref{fig:Failure Case III.B}, \ref{fig:Failure Case III.C}, and \ref{fig:Failure Case III.E} the planet is not found in the image, whereas in Fig.\ \ref{fig:second scenario} the planet is wrongly determined. In the figures, \color{green} $+$ \color{black} represents the real planet position projection, \color{blue} $\times$ \color{black} represents the expected planet position projection, and $\color{red} \square$ the found spikes, respectively.
\label{sec: test case}
\subsection{Filter Results}
\subsubsection{Navigation Concept of Operations}
\begin{figure}[H]
\centering
\includegraphics[width =\textwidth]{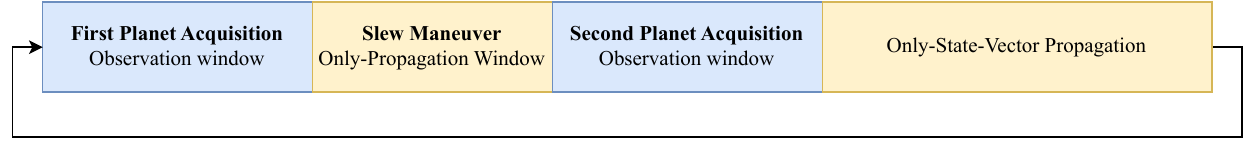}
\caption{Navigation Concept of Operations}
\label{fig:nav_cycle}
\end{figure}

In the study case, a CubeSat estimates its position and velocity by tracking visible planets over an interplanetary transfer. The spacecraft alternates observation windows, where an asynchronous tracking of the optimal pair of planets is performed, to only-propagation windows, where the filter only propagates the probe state as no external observations are acquired. The navigation CONOPS is shown in Fig. \ref{fig:nav_cycle}. The probe tracks the first planet of the optimal pair, which is selected at the beginning of the navigation cycle, then performs a slew maneuver to point to the second planet, during which no observations are acquired, and it observes this later. Eventually, the estimation is propagated until the following observation window starts.

At every time step of the planet observation, an image is generated using an improved version of the deep-space sky simulator in \cite{bella2021line}. The simulator models the effects caused by the lights, i.e., light-time and light-aberration effects, on the centroids' positions and by the impact of cosmic rays hitting the sensor frame. The sky simulator renders the image by taking into input the true probe pose and velocity. Since the attitude control system is not simulated in this work, the true probe orientation is computed by evaluating the desired pointing direction needed to acquire the planet at the center of the image and adding a random perturbation to it, which simulates the spacecraft jitter effect and the attitude knowledge error. %
Since the probe position is known with a given uncertainty (up to $10^5$ km in this work), the beacon projection will be not perfectly centered in the image, but still contained in it, which is a sufficient condition to let the IP pipeline extract the planet observation.

\subsubsection{Simulation Settings}
\begin{figure} [h!]
\centering
\includegraphics[width =0.8\textwidth]{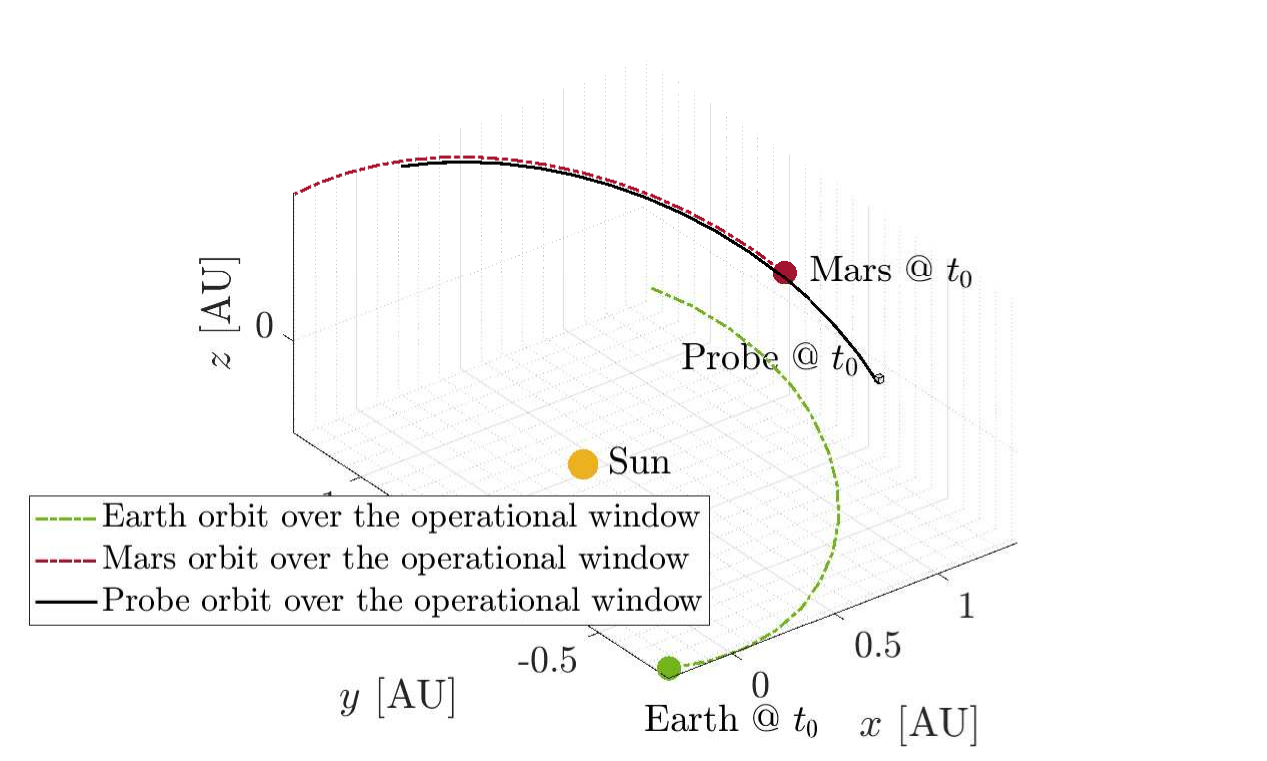}
\caption{Ballistic Interplanetary Reference Trajectory}
\label{fig:traj}
\end{figure}

The VBN filter proposed in this work is tested on an interplanetary high-fidelity ballistic trajectory between Earth--Mars \cite{merisio2022characterization}. The dynamics of the reference true trajectory include the SRP perturbations, the main attractor acceleration, third-body accelerations due to all the planets in the Solar System, and relativistic perturbations. Note that the dynamic model selected for the filter (Eq. \ref{eq:f}) is a lower-fidelity one, implying that the unmodeled accelerations are captured by the GM processes.
\cref{fig:traj} shows the analyzed leg of the nominal probe trajectory.

Starting from $t_0$, the estimation procedure begins. Each planet is tracked for an hour with a frequency of 0.01 Hz, the slew maneuver lasts 30 minutes, and the window in which the state is only propagated is ten days. Therefore, only two hours every ten days are reserved for correcting the state estimate. Over the interplanetary trajectory, 10 navigation legs of 10 days 2 hours, and 30 minutes each are repeated. 

\begin{figure}[h!]
\centering
\begin{subfigure}{0.45\textwidth}
\includegraphics[width =\textwidth]{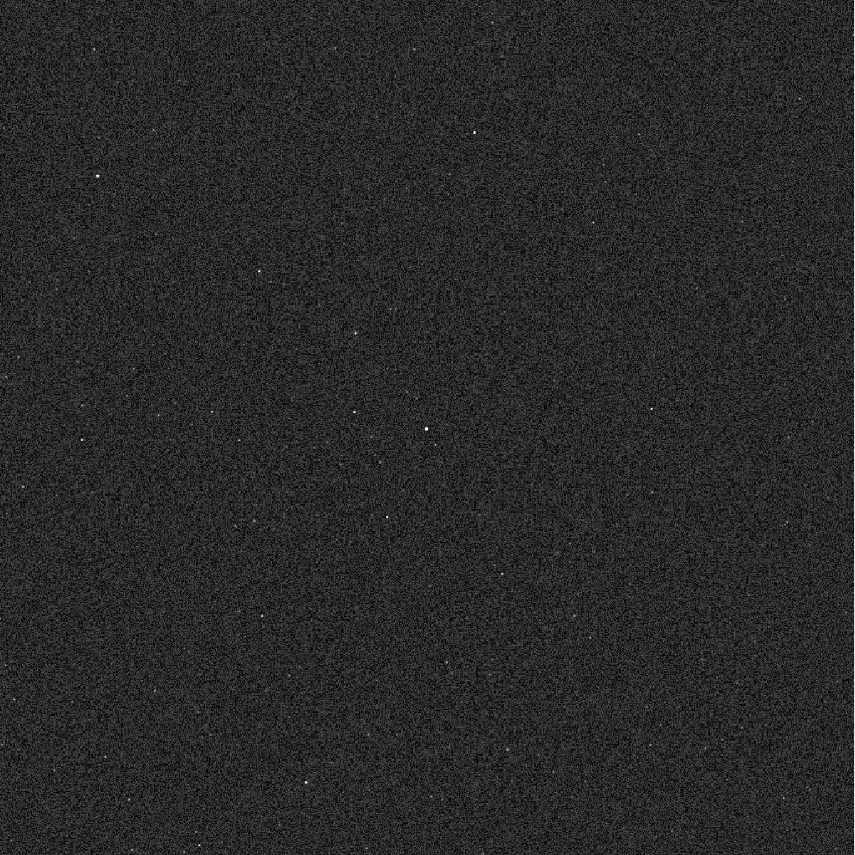}
\caption{Image taken during first planet observation}
\label{fig:nightsky Earth}
\end{subfigure}
\begin{subfigure}{0.45\textwidth}
\centering
\includegraphics[width = \textwidth]{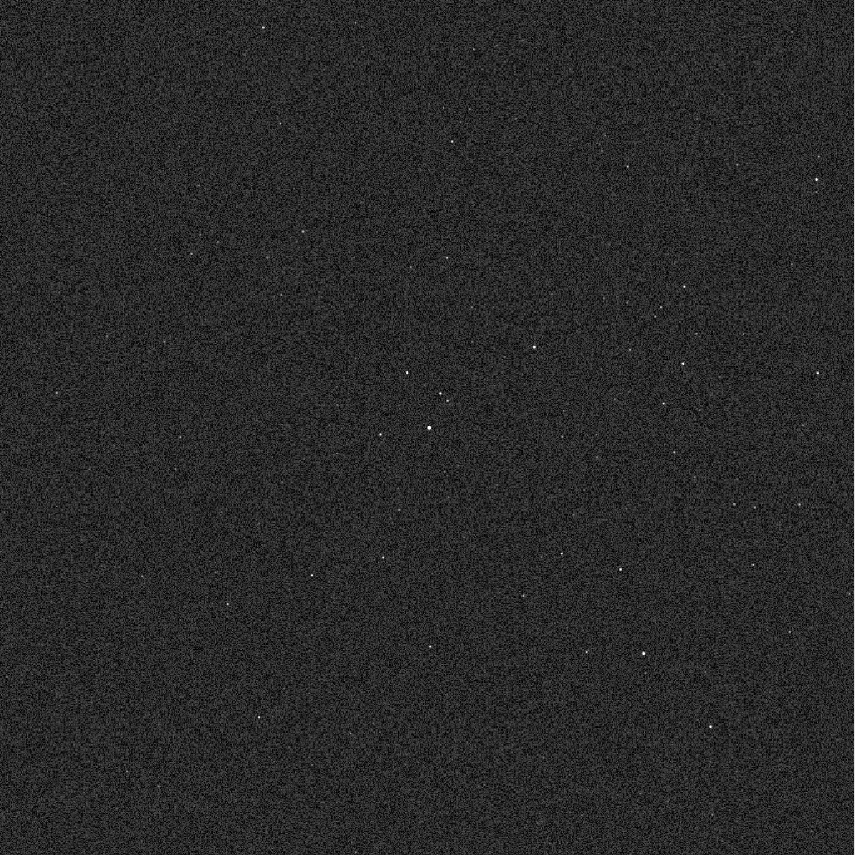}
\caption{Image taken during second planet observation}
\label{fig:nightsky mars}
\end{subfigure}
    \caption{Generated Sky-field Images}
\end{figure}

For image generation, the onboard camera is assumed to have the characteristics reported in Table~\ref{tab:camera}, where  F is the f-number, Q\textsubscript{e} is the quantum efficiency, T\textsubscript{lens} is the lens transmission, $\sigma_d$ is the defocus level, and $n_{\rm{CR}}$ is the number of single pixels that are turned on for simulating the presence of hitting cosmic rays.
\begin{table}[h!]
    \caption{Onboard Camera Characteristics.}
    \centering\begin{threeparttable}
    \begin{tabular}{c|c|c|c|c|c|c|c|c}
        \toprule\toprule
          FoV [deg] & F [-] & T [ms] & Image size [px] & f [mm] & Q\textsubscript{e}$\times$ T\textsubscript{lens} &  $\sigma_d$ [px] & $n_{\rm{CR}}$ & SAE\\
          20  &  2.2 & 400 & 1024$\times$ 1024 & 40 & 0.49 & 0.5 & 1 & 20\\
         \bottomrule \bottomrule
    \end{tabular}
    \end{threeparttable}
    \label{tab:camera}
\end{table}
Figs. \ref{fig:nightsky Earth} and \ref{fig:nightsky mars} are two of the rendered deep-space sky images adopted in the filtering procedure.
The initial standard deviations of the state adopted in Eq. \ref{eq:P} are reported in \cref{tab:initial_standard_deviation}.
\begin{table}[H]
    \centering   
    \caption[Accuracy of the state components at $t_0$]{Accuracy of the state components at $t_0$ }
     \begin{tabular}{cccc}
     \toprule \toprule
          $\sigma_{\textrm{r}}$ [km] &  $\sigma_{\textrm{v}}$  [km/s]  &  $\sigma_{\textrm{SRP}}$ [km/$s^{2}$]  &  $\sigma_{\textrm{R}}$ [km/$s^{2}$]  \\
\midrule
          $10^4$  & $10^{-1}$ & $10^{-10}$ & $10^{-10} $\\
          \bottomrule \bottomrule
    \end{tabular}
    \label{tab:initial_standard_deviation}
    \end{table}
\color{black}
Note that the values are selected following a conservative approach, taking into account that in deep space the initial position and velocity are usually known with an accuracy better than $10^4$ km and $0.1$ km/s, respectively. However, even if the performance of the IP procedure degrades by increasing the probe position uncertainty, this has been tested to work up to $\sigma_r = 10^6$ km with a success rate higher than 90\% in planet detection. Then, the performance of the IP worsens to 70\% when $\sigma_r = 10^7$ km (see Sec.\ \ref{sec:image processing results}) under the assumption of Gaussian errors in the measurement. For what concerns the OD, in \cite{andreis2022onboard} the algorithm has been tested to work up to $\sigma_r = 10^7$ km. Over this value, the OD algorithm is not able to select the optimal targets. 
Moreover, the standard deviation of the measurement error is set to $\sigma_{\rm{str}} = 0.1$ px, considering the results of the Monte Carlo runs in the extraction of the planet centroid reported in Sec.\ \ref{sec:image processing results}.
Eventually, only planets whose apparent magnitude is lower than 7 and whose SAE is greater than 20° are assumed to be visible by the camera, therefore, they are the only ones considered available for the optimal beacon selection process.
\subsubsection{Filter Performance} 
\begin{figure}[h!]
\centering
\includegraphics[width =\textwidth]{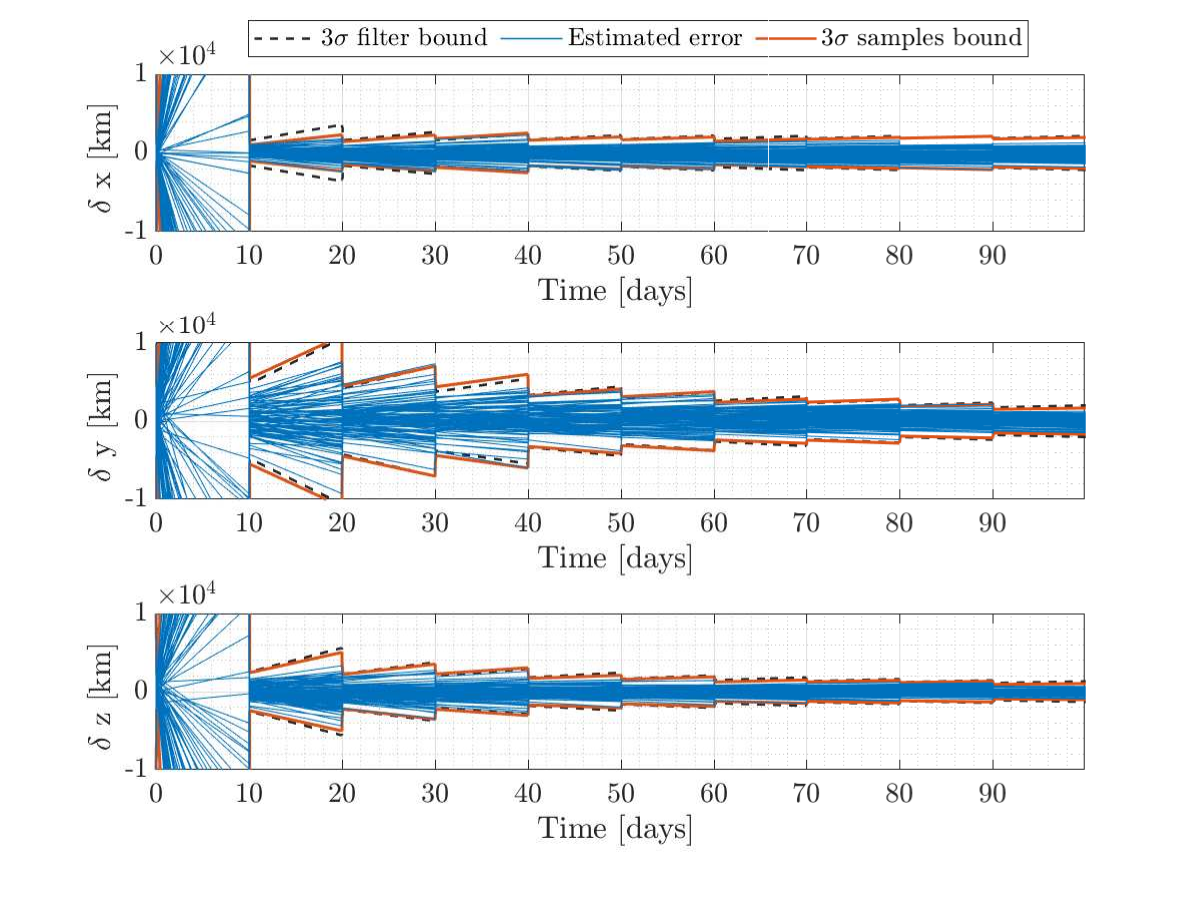}
\caption{Estimated errors for each position component with related $3\sigma$ bounds.}
\label{fig:pos_correction_IP}
\end{figure}

\begin{figure}[h!]
\centering
\includegraphics[width =\textwidth]{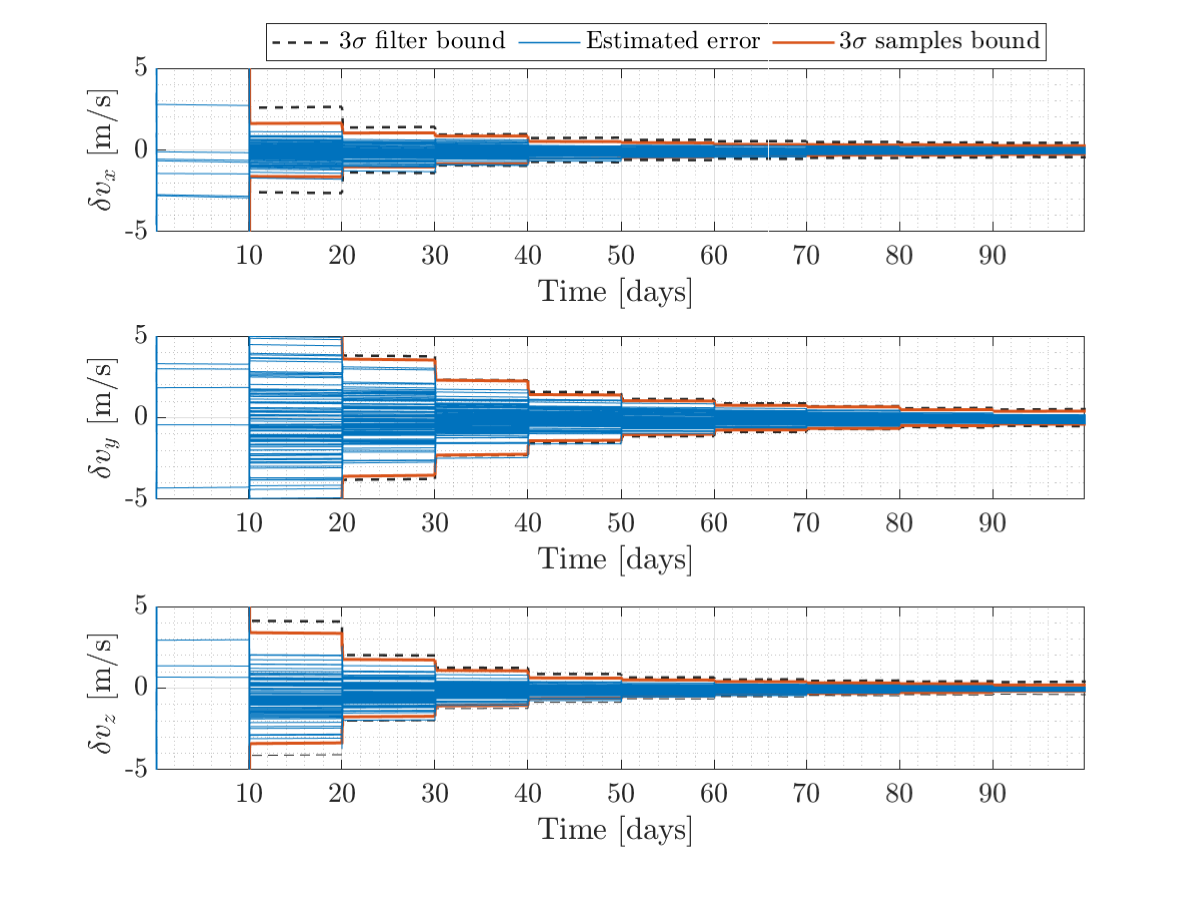}
\caption{Estimated errors for each velocity component with related $3\sigma$ bounds}
\label{fig:vel_correction_IP}
\end{figure}

A Monte Carlo simulation of 50 samples is performed. The initial probe state vector is perturbed by applying the $3\sigma$ standard deviation rule:
\begin{equation}
\bm x_0 = \tilde{\bm x}_0 + 3  \sqrt{\bm P_0}\bm k
\label{eq:x0}
\end{equation}
where $\tilde{\bm x}_0$ is the probe nominal state, $\bm k$ is a random vector with values within $[-1;1]$, and the square root operates on the elements of the initial error covariance matrix $\bm P_0$, which is defined as:
\begin{equation} \bm P_0 =
\textrm{diag}(\sigma_{\textrm{ r}}^2 \bm I_{3x3}, \sigma_{\textrm{v}}^2 \bm I_{3x3}, \sigma_{\textrm{R}}^2 \bm I_{3x3}, \sigma_{\textrm{SRP}}^2 \bm I_{3x3})
    \label{eq:P}
\end{equation}
\color{black}
The performance analyzed hereafter is relative to case 1 outlined in Sec.\ \ref{sec:measurement model}. Figures \ref{fig:pos_correction_IP} and \ref{fig:vel_correction_IP} show the position and velocity error profiles and $3\sigma$  covariance bounds in the J2000 ecliptic reference frame on the studied trajectory leg. The sample error profile is displayed with blue solid lines, whereas the orange solid lines and the dashed ones define the $3\sigma$ covariance bounds of the samples and the filter,
respectively.
At the end of the trajectory leg, the filter estimates the spacecraft position and velocity with a $3\sigma$ accuracy of 1953 km and 0.41 m/s, respectively. The 3$\sigma$ sample and filter covariance profiles are mostly overlapped, which suggests that the filter and its covariance matrices, in particular $\bm R$, are well tuned. This underlines that the planet centroids are extracted with a 3$\sigma$ accuracy lower than 0.3 px as found in the Monte Carlo campaigns conducted in Sec.\ \ref{sec:image processing results}. The outlier detection method has not rejected any planet determined by the IP. This is consistent with the results found in Sec.\ \ref{sec:image processing results}, where the percentage of false positives is lower the 1\% when the attitude is correctly determined and $\sigma_r$ is below $10^5$ km.
The planets observed during the interplanetary transfer are Earth, Mars, and Venus. Their object-to-pixel ratio is checked to be below 1 over the entire tracking period to respect the assumption of navigation with unresolved planets.
The performance of the filter detailed above (case 1) is compared with the other 4 cases:
\begin{itemize}
    \item Case 2: when the light-aberration effect on the planet position is corrected inside the IP (Sec.\ \ref{sec:light-aberration correction}).
    \item Case 3: when both effects are not corrected
    \item Case 4: when only the light-aberration effect is compensated
    \item Case 5: when only the light-time effect is compensated
    \end{itemize}
The root-mean-square error (RMSE) is chosen as the performance index to measure the estimation accuracy of the different models for the selected dataset. The latter comprises 50 initial state vectors and 814 images for simulation representing the scenarios observed during the interplanetary transfer. The filter and camera settings are unchanged with respect to case 1 for the sake of comparison.
\begin{figure}[h!]
\centering
\includegraphics[width =\textwidth]{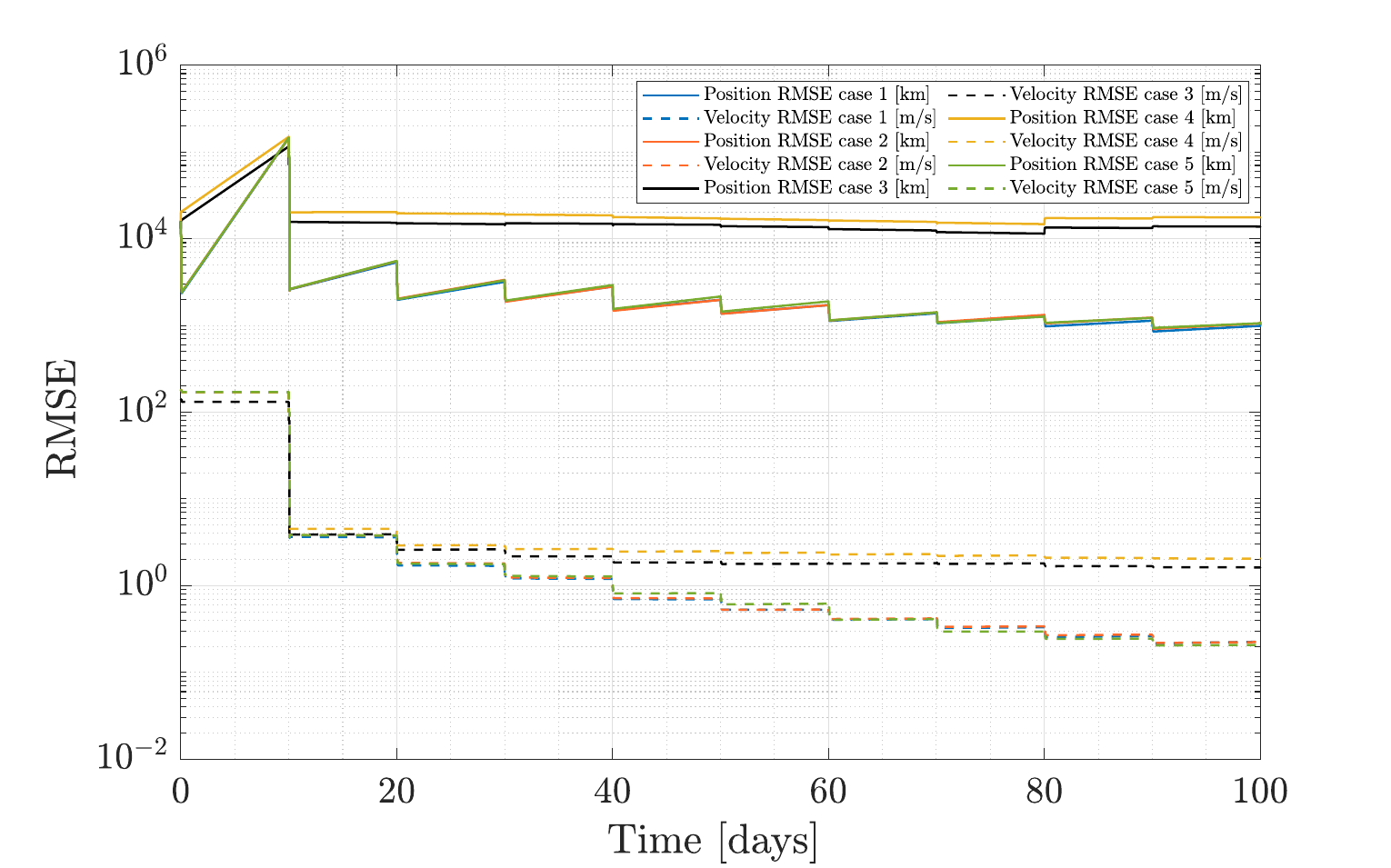}
\caption{Position and Velocity RMSE for the five cases.}
\label{fig:}
\end{figure}
The estimation is biased when the corrections of the light effects are not taken into account in the filter (case 3). Between the two effects, the greater deviation from the nominal value is obtained when the light-time effect is not compensated (case 4).  It can be noted that the RMSE associated with case 4 is greater than the one of case 3. This means that the deviation caused by the light-time effect is partially compensated by the light aberration. When the light-aberration correction is not applied on both the stars and planets centroids, the probe state estimation is affected by a small deviation (case 5). This occurs because the measurement model and the external observation are both affected by the light-aberration effect. Indeed, on one side, the attitude adopted to evaluate the measurement model is determined from stars in the image whose positions are aberrated. So, the measurement model is perturbed as well. On the other side, the external observation, i.e., the planet centroids, is not corrected for the light-aberration effect. Thus, the observation and the model are coherent.

Whereas, the performance of the filter when the light effects are both compensated in the measurement model (case 1) is analogous to the performance of the filter when the light aberration is corrected in the IP procedure (case 2). A slight discrepancy exists since the methodology followed to compensate for the light effects is different. In case 2, the light-aberration effect on the planet position is taken into account by correcting the observation, instead, in case 1, the light-aberration effect is compensated in the measurement model. Moreover, in case 1, the light-aberration effect is also taken into account in the derivation of $\bm H$ (Eq.\ \ref{eq:jacMeas}). Nevertheless, the similarity of the performance allows us to validate the innovative measurement model proposed in Sec.\ \ref{sec:measurement model} that provides an analytical first-order approximation of light effects corrections. 
\section{Conclusion}
This paper develops an autonomous vision-based navigation algorithm for interplanetary transfer with an application for CubeSats missions. A non-dimensional extended Kalman filter adopting the planet position projection as external observation is chosen by considering the limited processing capabilities of a standard miniaturized processor. Moreover, the measurements exploited for the estimation correction are directly extracted from images generated with a deep-space rendering engine. This procedure allows for obtaining a more faithful value of the measurement error and its influence on the filter solution.
At the end of the Earth--Mars trajectory, the filter estimates the spacecraft position and velocity with an accuracy of 2000 km and 0.5 m/s, respectively. Future analysis should examine the performance of the filter over low-thrust trajectories, which are desirable in CubeSats applications. Moreover, in this work, the probe attitude is determined from deep-space images with state-of-the-art star identification and attitude determination algorithms. The integration of the attitude filter in the proposed vision-based navigation algorithm and the adoption of a more robust star identification technique are objects of future investigation. Eventually, to validate the vision-based navigation applicability, it will be tested through hardware-in-the-loop simulations. Preliminary results have been already obtained through, first, the validation of the orbit determination algorithm with processor-in-the-loop simulations \cite{andreis2022onboard} and, second, with the validation of the image processing pipeline with the optical facility in the loop \cite{panicucci2022tinyv3rse,pugliatti2022tinyv3rse}. %

\section*{Acknowledgments}
This research is part of EXTREMA, a project that has received funding from the European Research Council (ERC) under the European Union’s Horizon 2020 research and innovation programme (Grant Agreement No.\ 864697). 

\bibliography{sample}

\begin{thebibliography}{46}
\newcommand{\enquote}[1]{``#1''}
\providecommand{\natexlab}[1]{#1}
\providecommand{\url}[1]{\texttt{#1}}
\providecommand{\urlprefix}{URL }
\expandafter\ifx\csname urlstyle\endcsname\relax
  \providecommand{\doi}[1]{\discretionary{}{}{}https://doi.org/#1}\else
  \providecommand{\doi}[1]{\discretionary{}{}{}\urlstyle{rm}\url{https://doi.org/#1}}\fi

\bibitem[{Wang et~al.(July 2014)Wang, Zheng, Sun, and Li}]{WANG201427}
Wang, Y., Zheng, W., Sun, S., and Li, L., \enquote{X-ray Pulsar-Based
  Navigation Using Time-Differenced Measurement,} \emph{Aerospace Science and
  Technology}, Vol.~36, July 2014, pp. 27--35.
\newblock \doi{10.1016/j.ast.2014.03.007}.

\bibitem[{Malgarini et~al.(2023)Malgarini, Franzese, and
  Topputo}]{malgarini2023}
Malgarini, A., Franzese, V., and Topputo, F., \enquote{Application of
  Pulsar-Based Navigation for Deep-Space CubeSats,} \emph{Aerospace}, Vol.~10,
  No.~8, 2023, p. 695.
\newblock \doi{10.3390/aerospace10080695}.

\bibitem[{Thornton and Border(2003)}]{oneway}
Thornton, C.~L., and Border, J.~S., \emph{Future Directions in Radiometric
  Tracking}, 1\textsuperscript{st} ed., John Wiley \& Sons, Ltd, New Jersey,
  United States, 2003, Chap.~5, pp. 63--75.
\newblock \doi{10.1002/0471728454.ch5}.

\bibitem[{Henry and Christian(2023)}]{henry2023absolute}
Henry, S., and Christian, J.~A., \enquote{Absolute Triangulation Algorithms for
  Space Exploration,} \emph{Journal of Guidance, Control, and Dynamics},
  Vol.~46, No.~1, 2023, pp. 21--46.
\newblock \doi{10.2514/1.G006989}.

\bibitem[{Maass et~al.(2020)Maass, Woicke, Oliveira, Razgus, and
  Kr{\"u}ger}]{maass2020crater}
Maass, B., Woicke, S., Oliveira, W.~M., Razgus, B., and Kr{\"u}ger, H.,
  \enquote{Crater Navigation System for Autonomous Precision Landing on the
  Moon,} \emph{Journal of Guidance, Control, and Dynamics}, Vol.~43, No.~8,
  2020, pp. 1414--1431.
\newblock \doi{10.2514/1.G004850}.

\bibitem[{Turan et~al.(April 2022)Turan, Speretta, and
  Gill}]{turan2022autonomous}
Turan, E., Speretta, S., and Gill, E., \enquote{Autonomous Navigation for
  Deep-Space Small Satellites: Scientific and Technological Advances,}
  \emph{Acta Astronautica}, Vol. 193, April 2022, pp. 56--74.
\newblock \doi{10.1016/j.actaastro.2021.12.030}.

\bibitem[{Bhaskaran et~al.(August 2000, No. AIAA-2000-3935)Bhaskaran, Riedel,
  Synnott, and Wang}]{bhaskaran2000deep}
Bhaskaran, S., Riedel, J., Synnott, S., and Wang, T., \enquote{The Deep Space 1
  Autonomous Navigation System-A Post-Flight Analysis,} \emph{Astrodynamics
  Specialist Conference}, AIAA, Denver, CO, USA, August 2000, No.
  AIAA-2000-3935.
\newblock \doi{10.2514/6.2000-3935}.

\bibitem[{Franzese et~al.(2021)Franzese, Topputo, Ankersen, and
  Walker}]{franzese2021deep}
Franzese, V., Topputo, F., Ankersen, F., and Walker, R., \enquote{Deep-Space
  Optical Navigation for M-ARGO Mission,} \emph{The Journal of the
  Astronautical Sciences}, Vol.~68, No.~4, 2021, pp. 1034--1055.
\newblock \doi{10.1007/s40295-021-00286-9}.

\bibitem[{Andreis et~al.(2022)Andreis, Franzese, and
  Topputo}]{andreis2022onboard}
Andreis, E., Franzese, V., and Topputo, F., \enquote{Onboard Orbit
  Determination for Deep-Space CubeSats,} \emph{Journal of Guidance, Control,
  and Dynamics}, Vol.~45, No.~8, 2022, pp. 1466--1480.
\newblock \doi{10.2514/1.G006294}.

\bibitem[{Merisio and Topputo(January 2023)}]{MERISIO2023115180}
Merisio, G., and Topputo, F., \enquote{Present-Day Model of Lunar Meteoroids
  and Their Impact Flashes for LUMIO Mission,} \emph{Icarus}, Vol. 389, January
  2023, p. 115180.
\newblock \doi{10.1016/j.icarus.2022.115180}.

\bibitem[{Panicucci et~al.(2023)Panicucci, Lebreton, Brochard, Zenou, and
  Delpech}]{panicucci2023shadow}
Panicucci, P., Lebreton, J., Brochard, R., Zenou, E., and Delpech, M.,
  \enquote{Shadow-Robust Silhouette Reconstruction for Small-Body
  Applications,} \emph{Journal of Spacecraft and Rockets}, Vol.~60, No.~3,
  2023, pp. 812--828.
\newblock \doi{https://doi.org/10.2514/1.A35444}.

\bibitem[{Panicucci et~al.(December 2023)Panicucci, Lebreton, Brochard, Zenou,
  and Delpech}]{panicucci2023vision}
Panicucci, P., Lebreton, J., Brochard, R., Zenou, E., and Delpech, M.,
  \enquote{Vision-based estimation of small body rotational state,} \emph{Acta
  Astronautica}, Vol. 213, December 2023, pp. 177--196.
\newblock \doi{10.1016/j.actaastro.2023.08.046}.

\bibitem[{Pugliatti et~al.(2022)Pugliatti, Franzese, and
  Topputo}]{pugliatti2022data}
Pugliatti, M., Franzese, V., and Topputo, F., \enquote{Data-Driven Image
  Processing for Onboard Optical Navigation Around a Binary Asteroid,}
  \emph{Journal of Spacecraft and Rockets}, Vol.~59, No.~3, 2022, pp. 943--959.
\newblock \doi{10.2514/1.A35213}.

\bibitem[{McCarthy et~al.(January 2022)McCarthy, Adam, Leonard, Antresian,
  Nelson, Sahr, Pelgrift, Lessac-Chenen, Geeraert, and Lauretta}]{leilah2022}
McCarthy, L.~K., Adam, C.~D., Leonard, J.~M., Antresian, P.~G., Nelson, D.,
  Sahr, E., Pelgrift, J., Lessac-Chenen, E.~J., Geeraert, J., and Lauretta, D.,
  \enquote{OSIRIS-REx Landmark Optical Navigation Performance During Orbital
  and Close Proximity Operations at Asteroid Bennu,} \emph{AIAA SCITECH 2022
  Forum}, AIAA, San Diego, CA \& Virtual, January 2022.
\newblock \doi{10.2514/6.2022-2520}.

\bibitem[{Leroy et~al.(2001)Leroy, Medioni, Johnson, and
  Matthies}]{LEROY2001787}
Leroy, B., Medioni, G., Johnson, E., and Matthies, L., \enquote{Crater
  detection for autonomous landing on asteroids,} \emph{Image and Vision
  Computing}, Vol.~19, No.~11, 2001, pp. 787--792.
\newblock \doi{10.1016/S0262-8856(00)00111-6}.

\bibitem[{Norman et~al.(2022)Norman, Miller, Olds, Mario, Palmer, Barnouin,
  Daly, Weirich, Seabrook, Bennett et~al.}]{norman2022autonomous}
Norman, C., Miller, C., Olds, R., Mario, C., Palmer, E., Barnouin, O., Daly,
  M., Weirich, J., Seabrook, J., Bennett, C., et~al., \enquote{Autonomous
  Navigation Performance Using Natural Feature Tracking during the OSIRIS-REx
  Touch-and-Go Sample Collection Event,} \emph{The Planetary Science Journal},
  Vol.~3, No.~5, 2022, p. 101.
\newblock \doi{10.3847/PSJ/ac5183}.

\bibitem[{Bhaskaran(June 2012, No.
  AIAA-2012-1267135)}]{bhaskaran2012autonomous}
Bhaskaran, S., \enquote{Autonomous Navigation for Deep Space Missions,}
  \emph{SpaceOps 2012}, AIAA, Stockholm, Sweden, June 2012, No.
  AIAA-2012-1267135.
\newblock \doi{10.2514/6.2012-1267135}.

\bibitem[{Topputo et~al.(2021)Topputo, Wang, Giordano, Franzese, Goldberg,
  Perez-Lissi, and Walker}]{topputo2021envelop}
Topputo, F., Wang, Y., Giordano, C., Franzese, V., Goldberg, H., Perez-Lissi,
  F., and Walker, R., \enquote{Envelop of Reachable Asteroids by M-ARGO
  CubeSat,} \emph{Advances in Space Research}, Vol.~67, No.~12, 2021, pp.
  4193--4221.
\newblock \doi{10.1016/j.asr.2021.02.031}.

\bibitem[{Karimi and Mortari(2015)}]{raymond2015interplanetary}
Karimi, R.~R., and Mortari, D., \enquote{Interplanetary Autonomous Navigation
  Using Visible Planets,} \emph{Journal of Guidance, Control, and Dynamics},
  Vol.~38, No.~6, 2015, pp. 1151--1156.
\newblock \doi{10.2514/1.G000575}.

\bibitem[{Casini et~al.(2023)Casini, Cervone, Monna, and Gill}]{casini2022line}
Casini, S., Cervone, A., Monna, B., and Gill, E., \enquote{On line-of-sight
  navigation for deep-space applications: A performance analysis,}
  \emph{Advances in Space Research}, Vol.~72, No.~7, 2023, pp. 2994--3008.
\newblock \doi{10.1016/j.asr.2022.12.017}.

\bibitem[{Stastny and Geller(2008)}]{stastny2008autonomous}
Stastny, N.~B., and Geller, D.~K., \enquote{Autonomous Optical Navigation at
  Jupiter: a Linear Covariance Analysis,} \emph{Journal of Spacecraft and
  Rockets}, Vol.~45, No.~2, 2008, pp. 290--298.
\newblock \doi{10.2514/1.28451}.

\bibitem[{Bhaskaran(AIAA, Virtual Event, November 2020,
  AIAA-2020-4139)}]{bhaskaran2020autonomous}
Bhaskaran, S., \enquote{Autonomous Optical-Only Navigation for Deep Space
  Missions,} \emph{ASCEND 2020}, AIAA, Virtual Event, November 2020,
  AIAA-2020-4139.
\newblock \doi{10.2514/6.2020-4139}.

\bibitem[{Vaughan et~al.(August 1992, No.\ AIAA-92-4522-CP)Vaughan, Riedel,
  Davis, OWEN, and Synnott}]{vaughan1992optical}
Vaughan, R., Riedel, J., Davis, R., OWEN, W., JR, and Synnott, S.,
  \enquote{Optical Navigation for the Galileo Gaspra Encounter,}
  \emph{Astrodynamics Conference}, AIAA, Hilton Head Island, SC, U.S.A., August
  1992, No.\ AIAA-92-4522-CP.
\newblock \doi{10.2514/6.1992-4522}.

\bibitem[{Andreis et~al.(2024)Andreis, Panicucci, Franzese, and
  Topputo}]{andreis2022robust}
Andreis, E., Panicucci, P., Franzese, V., and Topputo, F., \enquote{A Robust
  Image Processing Pipeline for Planets Line-Of-Sight Extraction for Deep-Space
  Autonomous Cubesats Navigation,} \emph{Proceedings of the 44th Annual
  American Astronautical Society Guidance, Navigation, and Control Conference,
  2022}, Springer International Publishing, New York City, United States, 2024,
  pp. 1103--1121.
\newblock \doi{10.1007/978-3-031-51928-4_61}.

\bibitem[{Broschart et~al.(2019)Broschart, Bradley, and Bhaskaran}]{kinematic}
Broschart, S.~B., Bradley, N., and Bhaskaran, S., \enquote{Kinematic
  Approximation of Position Accuracy Achieved Using Optical Observations of
  Distant Asteroids,} \emph{Journal of Spacecraft and Rockets}, Vol.~56, No.~5,
  2019, pp. 1383--1392.
\newblock \doi{10.2514/1.A34354}.

\bibitem[{Franzese and Topputo(2023)}]{franzese2023celestial}
Franzese, V., and Topputo, F., \enquote{Celestial Bodies Far-Range Detection
  with Deep-Space CubeSats,} \emph{Sensors}, Vol.~23, No.~9, 2023, p. 4544.
\newblock \doi{https://doi.org/10.3390/s23094544}.

\bibitem[{Franzese and Topputo(2020)}]{franzese2020optimal}
Franzese, V., and Topputo, F., \enquote{Optimal Beacons Selection for
  Deep-Space Optical Navigation,} \emph{The Journal of the Astronautical
  Sciences}, Vol.~67, No.~4, 2020, pp. 1775--1792.
\newblock \doi{10.1007/s40295-020-00242-z}.

\bibitem[{Kazemi et~al.(2015)Kazemi, Enright, and Dzamba}]{kazemi2015improving}
Kazemi, L., Enright, J., and Dzamba, T., \enquote{Improving Star Tracker
  Centroiding Performance in Dynamic Imaging Conditions,} \emph{2015 IEEE
  Aerospace Conference}, IEEE, Big Sky, MT, USA, 2015, pp. 1--8.
\newblock \doi{10.1109/AERO.2015.7119226}.

\bibitem[{Wan et~al.(2018)Wan, Wang, Wei, Li, and Zhang}]{wan2018star}
Wan, X., Wang, G., Wei, X., Li, J., and Zhang, G., \enquote{Star Centroiding
  Based on Dast Gaussian Fitting for Star Sensors,} \emph{Sensors}, Vol.~18,
  No.~9, 2018, p. 2836.
\newblock \doi{10.3390/s18092836}.

\bibitem[{Mortari(1997)}]{search-less}
Mortari, D., \enquote{Search-Less Algorithm for Star Pattern Recognition,}
  \emph{Journal of the Astronautical Sciences}, Vol.~45, 1997, p. 179–194.
\newblock \doi{10.1007/BF03546375}.

\bibitem[{Bentley and Sedgewick(1997)}]{bentley1997fast}
Bentley, J.~L., and Sedgewick, R., \enquote{Fast algorithms for sorting and
  searching strings,} \emph{Proceedings of the Eighth Annual ACM-SIAM Symposium
  on Discrete Algorithms}, Society for Industrial and Applied Mathematics,
  Philadelphia, PA, USA, 1997, p. 360–369.

\bibitem[{Mortari and Neta(2000)}]{mortari2014k}
Mortari, D., and Neta, B., \enquote{K-vector range searching techniques,}
  \emph{Adv. Astronaut. Sci}, Vol. 105, No.~1, 2000, pp. 449--464.

\bibitem[{Markley and Crassidis(2014)}]{markley2014fundamentals}
Markley, F.~L., and Crassidis, J.~L., \emph{Fundamentals of Spacecraft Attitude
  Determination and Control}, 1\textsuperscript{st} ed., Springer, New York,
  2014, Chap.~5, pp. 183--235.
\newblock \doi{10.1007/978-1-4939-0802-8}.

\bibitem[{Fischler and Bolles(1981)}]{fischler1981random}
Fischler, M.~A., and Bolles, R.~C., \enquote{Random Sample Consensus: A
  Paradigm for Model Fitting with Applications to Image Analysis and Automated
  Cartography,} \emph{Communications of the ACM}, Vol.~24, No.~6, 1981, pp.
  381--395.
\newblock \doi{10.1145/358669.358692}.

\bibitem[{Hartley and Zisserman(2004)}]{ransac}
Hartley, R., and Zisserman, A., \emph{Multiple View Geometry in Computer
  Vision}, 2\textsuperscript{nd} ed., Cambridge University Press 2000, UK,
  2004, Chap.~3, pp. 117--120.

\bibitem[{Carpenter and D’Souza(2018)}]{bestpractise}
Carpenter, J.~R., and D’Souza, C.~N., \enquote{{Navigation Filter Best
  Practices},} Tech. Rep. 20180003657, NASA, 04 2018.

\bibitem[{Jean et~al.(December 2019)Jean, Ng, and Misra}]{cannonball_model}
Jean, I., Ng, A., and Misra, A.~K., \enquote{Impact of Solar Radiation Pressure
  Modeling on Orbital Dynamics in the Vicinity of Binary Asteroids,} \emph{Acta
  Astronautica}, Vol. 165, December 2019, pp. 167--183.
\newblock \doi{10.1016/j.actaastro.2019.09.003}.

\bibitem[{Mortari and Conway(2017)}]{mortari2017single}
Mortari, D., and Conway, D., \enquote{Single-Point Position Estimation in
  Interplanetary Trajectories Using Star Trackers,} \emph{Celestial Mechanics
  and Dynamical Astronomy}, Vol. 128, No.~1, 2017, pp. 115--130.
\newblock \doi{10.1007/s10569-016-9738-4}.

\bibitem[{Shuster(August 2003)}]{shuster2003stellar}
Shuster, M.~D., \enquote{Stellar Aberration and Parallax: A Tutorial,}
  \emph{The Journal of the astronautical sciences}, Vol.~51, August 2003, pp.
  477--494.
\newblock \doi{10.1007/BF03546295}.

\bibitem[{Christian(2019)}]{christian2019starnav}
Christian, J.~A., \enquote{StarNAV: Autonomous Optical Navigation of a
  Spacecraft by the Relativistic Perturbation of Starlight,} \emph{Sensors},
  Vol.~19, No.~19, 2019, p. 4064.
\newblock \doi{10.3390/s19194064}.

\bibitem[{Liu et~al.(2004)Liu, Shah, and Jiang}]{liu2004line}
Liu, H., Shah, S., and Jiang, W., \enquote{On-line Outlier Detection and Data
  Cleaning,} \emph{Computers \& chemical engineering}, Vol.~28, No.~9, 2004,
  pp. 1635--1647.
\newblock \doi{10.1016/j.compchemeng.2004.01.009}.

\bibitem[{Mortari et~al.(2004)Mortari, Samaan, Bruccoleri, and
  Junkins}]{mortari2004pyramid}
Mortari, D., Samaan, M.~A., Bruccoleri, C., and Junkins, J.~L., \enquote{The
  Pyramid Star Identification Technique,} \emph{Navigation}, Vol.~51, No.~3,
  2004, pp. 171--183.
\newblock \doi{10.1002/j.2161-4296.2004.tb00349.x}.

\bibitem[{Bella et~al.(2022)Bella, Andreis, Franzese, Panicucci, Topputo
  et~al.}]{bella2021line}
Bella, S., Andreis, E., Franzese, V., Panicucci, P., Topputo, F., et~al.,
  \enquote{Line-of-Sight Extraction Algorithm for Deep-Space Autonomous
  Navigation,} \emph{Advances In The Astronautical Sciences}, Vol. 177, 2022,
  pp. 1--18.

\bibitem[{Merisio et~al.(2022, No.\ AAS 21-677)Merisio, Topputo
  et~al.}]{merisio2022characterization}
Merisio, G., Topputo, F., et~al., \enquote{Characterization of ballistic
  capture corridors aiming at autonomous ballistic capture at Mars,}
  \emph{Advances In The Astronautical Sciences}, Vol. 177, 2022, No.\ AAS
  21-677.

\bibitem[{Panicucci and Topputo(2022)}]{panicucci2022tinyv3rse}
Panicucci, P., and Topputo, F., \enquote{The TinyV3RSE Hardware-in-the-Loop
  Vision-Based Navigation Facility,} \emph{Sensors}, Vol.~22, No.~23, 2022, p.
  9333.
\newblock \doi{10.3390/s22239333}.

\bibitem[{Pugliatti et~al.(January 2022)Pugliatti, Franzese, Panicucci, and
  Topputo}]{pugliatti2022tinyv3rse}
Pugliatti, M., Franzese, V., Panicucci, P., and Topputo, F.,
  \enquote{TINYV3RSE: The DART Vision-Based Navigation Test-bench,} \emph{AIAA
  Scitech 2022 Forum}, AIAA, San Diego, CA \& Virtual, January 2022, p. 1193.
\newblock \doi{10.2514/6.2022-1193}.

\end{thebibliography}

\end{document}